\documentclass[journal]{IEEEtran}

\usepackage{cite}
\usepackage{graphicx}
\usepackage{amsmath,amssymb,amsfonts}
\usepackage{bm}
\usepackage{booktabs}
\usepackage[table]{xcolor}
\usepackage{array}
\usepackage{url}
\usepackage{microtype}
\usepackage[hidelinks]{hyperref}
\hypersetup{pdftitle={PhysSkin++: Generalizable Self-Supervised Neural Skinning for Real-Time Interactive Physics-Based Animation}}

\DeclareGraphicsExtensions{.pdf,.png,.jpg,.jpeg}

\definecolor{colorFst}{HTML}{F59194}
\definecolor{colorSnd}{HTML}{FAC791}
\definecolor{colorThd}{HTML}{FFFF99}
\definecolor{codexpurple}{HTML}{6A3FB5}
\newcommand{\codex}[1]{{\color{black}#1}}

\newcommand{\defX}{\bm{x}}
\newcommand{\refX}{\mathbf{X}}
\newcommand{\dofs}{\mathbf{z}}

\newcommand{\R}{\mathbb{R}}
\newcommand{\W}{\bm{W}}
\newcommand{\dofsMat}{\mathbf{Z}}

\newcommand{\mass}{\mathbf{M}}
\newcommand{\NumHandle}{m}
\newcommand{\Jacobian}{\bm{J}}

\newcommand{\DeformationMap}{\phi}

\newcommand{\ElasticPotential}{E_\textrm{pot}}
\newcommand{\ForceEnergy}{E_\textrm{force}}
\newcommand{\BoundaryEnergy}{E_\textrm{bnd}}
\newcommand{\CollisionEnergy}{E_\textrm{coll}}
\newcommand{\Domain}{\Omega}
\newcommand{\DirichletBoundary}{\Gamma_D}
\newcommand{\DirichletMap}{\phi_D}
\newcommand{\BoundaryValue}{\mathbf{g}}

\renewcommand{\eqref}[1]{Equation~\ref{eq:#1}}

\definecolor{myPurple}{rgb}{0.4, .0, .8}
\definecolor{myGreen}{rgb}{0, 0.6, .3}
\definecolor{myRed}{rgb}{0.8, .2, .2}
\definecolor{myOrange}{rgb}{0.8, 0.45, 0.0}
\definecolor{myBlue}{rgb}{.0, .0, 1.0}
\definecolor{myBlue2}{rgb}{.0, 1.0, 1.0}
\definecolor{myBlack}{rgb}{.0, .0, 0.0}
\definecolor{darkmidnightblue}{rgb}{0.0, 0.2, 0.4}
\definecolor{MyGreen}{rgb}{0.02,0.5,0.02}

\begin{document}

\title{{\fontsize{20}{24}\selectfont PhysSkin++: Generalizable Self-Supervised Neural Skinning\\for Real-Time Interactive Physics-Based Animation}}

\author{Yuanhang~Lei,
        Baofei~Liu,
        Xingxuan~Li,
        Tao~Cheng,
        Zixuan~Wang,
        Boming~Zhao,
        Yujun~Shen,
        Siyuan~Huang,
        Ruizhen~Hu,
        Peter~Yichen~Chen,
        Hujun~Bao,
        and~Zhaopeng~Cui\textsuperscript{$\dagger$}%
\thanks{$\dagger$ Corresponding author.}%
\thanks{Yuanhang Lei, Baofei Liu, Xingxuan Li, Tao Cheng, Zixuan Wang, Boming Zhao, Hujun Bao, and Zhaopeng Cui are with the State Key Laboratory of CAD\&CG, Zhejiang University.}%
\thanks{Yujun Shen is with Ant Group, Hangzhou 310000, China.}%
\thanks{Siyuan Huang is with the Beijing Institute for General Artificial Intelligence (BIGAI).}%
\thanks{Ruizhen Hu is with Shenzhen University.}%
\thanks{Peter Yichen Chen is with the University of British Columbia.}}

\pagestyle{plain}

\maketitle

\begin{abstract}
Generalizing real-time physics-based animation across diverse 3D shapes and spatial discretizations remains challenging. In this paper, we introduce PhysSkin++, a physics-informed framework that learns continuous neural skinning fields from static geometries to construct compact deformation subspaces. Specifically, we propose a Neural Skinning Fields Autoencoder that predicts shape-dependent deformation subspaces without per-object network training. To learn these subspaces without simulation trajectories, we develop a Physics-Informed Self-Supervised Learning strategy that combines skinning-field normalization with conflict-aware gradient correction, stabilizing joint optimization under physical and orthogonality constraints. The learned representation supports efficient reduced dynamics with user control and contact interactions, together with direct animation binding to meshes, point clouds, and 3D Gaussians. For prescribed rig animation, we further introduce a self-supervised complementarity objective to learn complementary elastic responses without enforcing complementarity constraints at runtime. Experiments demonstrate cross-shape generalization of the learned skinning fields, real-time reduced simulation, and applications to controllable, collision-aware, and rig-complementary animation. Project page: \url{https://zju3dv.github.io/PhysSkin_plus/}.
\end{abstract}

\begin{IEEEkeywords}
Physics-based animation, geometric deep learning, reduced-order modeling, self-supervised learning.
\end{IEEEkeywords}

\begin{figure*}[!t]
    \centering
    \includegraphics[width=\textwidth]{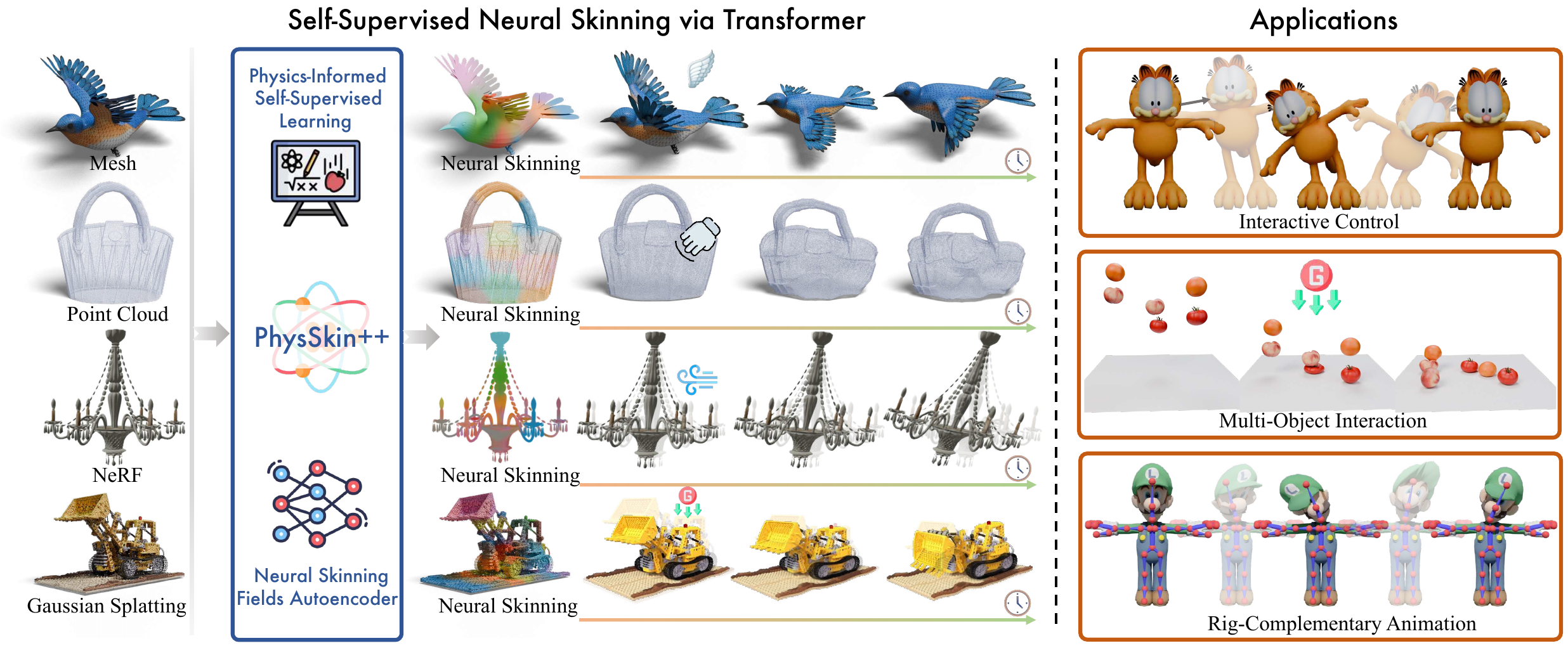}
    \caption{PhysSkin++ is a generalizable physics-informed neural skinning framework for object animation. The framework is learned directly from static 3D geometries via physics-informed self-supervision without any annotated data. Once trained, PhysSkin++ can be applied in a feed-forward manner to perform neural skinning for diverse 3D shapes and discretizations, enabling real-time physics-based animation. It further supports interactive control, collision-aware multi-object interaction, and rig-complementary animation.}
    \label{fig:teaser}
    \vspace{-1.1em}
\end{figure*}

\section{Introduction}
\label{sec:intro}

\noindent Real-time physics-based animation is a long-standing goal in computer vision and graphics, underpinning applications in VR/AR authoring, character animation, and interactive digital content creation. To support such applications, an animation system must provide a compact yet expressive motion representation that captures physically plausible deformations, accommodates user control and interactions, and remains computationally efficient for real-time animation.

Subspace physics-based animation~\cite{an2008optimizing, chen2023crom} addresses this by learning a low-dimensional motion subspace for efficient computation, and then lifting the result back to the full space using a subspace mapping. However, classical methods~\cite{an2008optimizing, sifakis2012fem} optimize a single linear subspace mapping matrix tied to a particular mesh topology and resolution, which prevents generalization across different spatial discretizations. Recently, neural methods~\cite{chen2023crom, sharp2023data} have emerged as promising alternatives by using neural networks to learn subspace mappings that naturally handle arbitrary spatial discretizations. These approaches, however, must still train a separate network per object and fail to generalize to different 3D shapes. In parallel, recent transformer-based approaches learn physical dynamics directly from simulation trajectories across diverse objects and physical systems~\cite{dou2026rigidformer,wang2026worldparticle}. Although this data-driven route enables generalizable neural simulation, it requires extensive trajectory supervision and does not explicitly construct a low-dimensional deformation subspace for reduced-order animation. These limitations motivate a central question: \emph{How can we learn a physics-consistent deformation subspace mapping that generalizes across diverse 3D shapes and discretizations directly from static geometries, without relying on supervised training data?}

In this work, we introduce PhysSkin++ (Fig.~\ref{fig:teaser}), a generalizable physics-informed framework for real-time physics-based animation across diverse 3D shapes and discretizations. In the spirit of Linear Blend Skinning (LBS)~\cite{magnenat1989joint}, we use continuous skinning fields to define a deformation subspace mapping, with handle transformations serving as the reduced coordinates for full-space 3D object deformation. To construct this representation directly from static geometries, we introduce a Neural Skinning Fields Autoencoder. Specifically, PhysSkin++ employs a transformer-based point cloud encoder to extract latent shape features from surface samples, while a cross-attention decoder predicts continuous neural skinning fields at arbitrary spatial locations. PhysSkin++ requires no simulation trajectories, skinning annotations, or category-specific priors. This mesh-free, discretization-agnostic representation generalizes across object categories, topologies, and resolutions.

Learning such a deformation subspace mapping purely from geometry is highly non-trivial. A physically meaningful subspace must exhibit low potential energy, reflecting deformation modes compatible with physically plausible behavior; spatial smoothness, ensuring coherent deformations; and orthogonality, promoting independence among the deformation modes. We therefore formulate a self-supervised objective consisting of potential-energy, smoothness, and orthogonality terms. However, jointly optimizing these terms often leads to conflicts in gradient magnitude and direction, making naive joint optimization unstable. To overcome this optimization challenge, we introduce a Physics-Informed Self-Supervised Learning (PISSL) strategy. First, we apply on-the-fly skinning-field normalization, which regulates the scale of the learned skinning weights and prevents numerical drift during training. Second, we incorporate a conflict-aware gradient correction mechanism~\cite{liu2025config} to resolve destructive interference among gradients of the loss terms, enabling balanced optimization and stable convergence. Together, these mechanisms enable PhysSkin++ to learn a physically consistent deformation subspace mapping directly from static geometries, with spatially smooth and approximately orthogonal skinning fields.

At runtime, we advance the reduced coordinates through implicit reduced dynamics and map them to full-space deformations through the learned skinning-based subspace mapping. To support controllable and collision-aware animation, we incorporate Dirichlet boundary conditions, interactive control, environmental interaction, and multi-object contact into this formulation. Prescribed boundary displacements are enforced through a boundary-compatible deformation map, while external forces and contact interactions are incorporated through their corresponding energy terms. This formulation supports both rigid--deformable and deformable--deformable interactions within the learned deformation subspaces. Moreover, the continuous deformation map can be directly bound to different spatial discretizations, including 3D Gaussian representations~\cite{kerbl20233d}, enabling physics-based animation of diverse object representations within the same reduced dynamics framework.

Beyond purely physics-driven animation, PhysSkin++ also supports rig-complementary animation when an artist-authored LBS rig is available. Artist-authored rigs provide intuitive low-dimensional control, but their kinematic deformations alone cannot reproduce physical elastic responses, while increasing the complexity of a rig places a greater burden on animation authoring. Complementary dynamics~\cite{Zhang:CompDynamics:2020,benchekroun2023fast} addresses this tension by augmenting a prescribed rig animation with complementary physical deformations. Following this principle, PhysSkin++ treats the user-authored rig-based animation as the primary animation and learns neural skinning fields for complementary elastic response. We augment the self-supervised objective with a rig-space complementarity term defined by the rig Jacobian, encouraging the learned deformation subspace to be complementary to the prescribed rig animation space. At runtime, the resulting reduced dynamics enriches the prescribed animation with physics-based complementary displacement without enforcing the rig-space complementarity constraint at every step. Both animation settings share the continuous skinning representation and reduced dynamics framework.

Experiments on various datasets evaluate the learned skinning fields and demonstrate real-time physics-based animation across diverse shapes, controls, interactions, and spatial discretizations. Our contributions are:
\begin{itemize}
    \item We present \textbf{PhysSkin++}, a generalizable physics-informed framework for real-time physics-based animation across diverse 3D shapes and discretizations by learning neural skinning fields directly from static geometries. Its Neural Skinning Fields Autoencoder employs a transformer-based shape encoder and a cross-attention decoder to predict continuous skinning fields.
    \item We develop a \textbf{Physics-Informed Self-Supervised Learning (PISSL) strategy} that incorporates on-the-fly skinning-field normalization and conflict-aware gradient correction, enabling stable joint optimization of multiple self-supervised loss terms.
    \item We develop a \textbf{controllable and collision-aware reduced dynamics formulation} that supports Dirichlet boundary conditions, interactive control, environmental interaction, and multi-object contact, together with discretization-agnostic binding of the continuous deformation map to meshes, point clouds, and 3D Gaussian representations.
    \item We support \textbf{rig-complementary animation} by extending the self-supervised objective with a rig-space complementarity term defined by the rig Jacobian. This enables physics-based complementary elastic responses under interactive rig control and user-authored animation sequences without enforcing the rig-space complementarity constraint at runtime.
\end{itemize}

A preliminary version of this work has been published as PhysSkin~\cite{lei2026physskin}. The present article extends the conference version in four aspects. First, we develop a reduced dynamics formulation for controllable and collision-aware animation, supporting prescribed boundary displacements, interactive control, environmental interaction, and multi-object contact. Second, we provide a discretization-agnostic binding formulation for 3D Gaussian representations, updating both Gaussian centers and anisotropic covariances through local affine transformations determined by the learned skinning fields and current reduced coordinates. Third, we extend the deformation representation with prescribed rig displacement and the self-supervised objective with a rig-space complementarity term. These extensions support complementary elastic deformation for prescribed rig animations without enforcing the complementarity constraint at runtime. Finally, we expand the experimental evaluation to cover controllable and collision-aware animation, 3D Gaussian animation, and self-supervised rig-complementarity learning, including rig-complementary animation under interactive rig control and user-authored animation sequences.

\section{Related Work}
\codex{\noindent\textbf{Neural Physics-Based Animation.}}
Subspace physics-based animation~\cite{an2008optimizing, von2013efficient, chen2023crom} is an effective method for accelerating animation computations. Its core idea is to learn a low-dimensional motion subspace for efficient computation, and then lift the result back to the full space using a subspace mapping. Recently, fully neural approaches have been proposed to directly learn subspace mapping by neural networks. Data-driven methods~\cite{fulton2019latent, shen2021high, chen2023crom, zong2023neural, chang2023licrom} represented by  CROM~\cite{chen2023crom} learn the mapping for a single object by fitting neural networks to motion sequences from physical simulators, but they have limited generalization ability to unseen motion patterns. \codex{In parallel, recent transformer-based approaches learn physical dynamics directly from simulation trajectories across rigid, deformable, and particle-based systems~\cite{dou2026rigidformer,wang2026worldparticle,chen2026physiformer,Geng_2026_CVPR}.} Data-free methods~\cite{sharp2023data, lyu2024accelerate, modi2024simplicits, wang2024neural, chang2025shape} learn a low-energy subspace mapping by physics-informed learning without requiring motion sequences, but pure neural network approaches often struggle to converge to high-quality subspace representations. Recently, Simplicits~\cite{modi2024simplicits} learns neural skinning weight functions as a physics-informed subspace mapping by minimizing a quadratic energy subject to orthogonality constraints, which stabilizes training compared to purely neural models but limits generalization due to its per-object network design. Our method overcomes these limitations with a neural skinning fields autoencoder that outputs skinning weights generalizing across diverse 3D shapes, enabling real-time subspace physics-based animation. \codex{Our physics-informed neural subspace formulation retains explicit reduced-order dynamics and complements approaches that learn dynamics directly from simulation trajectories.}

\noindent\textbf{Neural Rigging and Skinning Representations.}
Another classical reduced deformation representation is skeletal rigging-based animation~\cite{baran2007automatic, feng2015avatar, SMPL:2015}. Recently, based on 3D shape VAEs~\cite{zhang20233dshape2vecset, zhao2023michelangelo}, many works~\cite{deng2025anymate, song2025magicarticulate, guo2025make, liu2025riganything, xu2020rignet, zhang2025one, song2025puppeteer} have performed supervised learning from expert-annotated 3D object rigging and skinning datasets to infer objects' joints, bones, and skinning weights through a feed-forward network. However, some methods are limited to specific categories of 3D shapes such as characters~\cite{xu2020rignet, guo2025make} or animals~\cite{sun2025ponymation, wu2022casa}, as they incorporate category-specific priors~\cite{SMPL:2015, zuffi20173d} such as predefined skeletal structures or semantic constraints, making it difficult to generalize to other object categories. Works represented by Anymate~\cite{deng2025anymate} learn from annotated 3D object rigging and skinning datasets and design category-agnostic network architectures, thereby enabling the ability to generalize across diverse object categories and demonstrating stronger generalization capabilities. However, these methods heavily rely on expert-annotated data~\cite{deng2025anymate, li20214dcomplete, li2025puppet, wu2022casa, luo2023rabit}, which are costly and difficult to scale. In contrast, our work employs self-supervised learning to train neural skinning fields directly from static 3D geometries without any annotated data.

\raggedbottom
\setlength{\textfloatsep}{12pt}
\section{Method}

\subsection{Framework Overview and Problem Formulation}
\label{subsec:skinning_eigenmodes}
\begin{figure*}[t]
    \centering
    \includegraphics[width=0.98\linewidth]{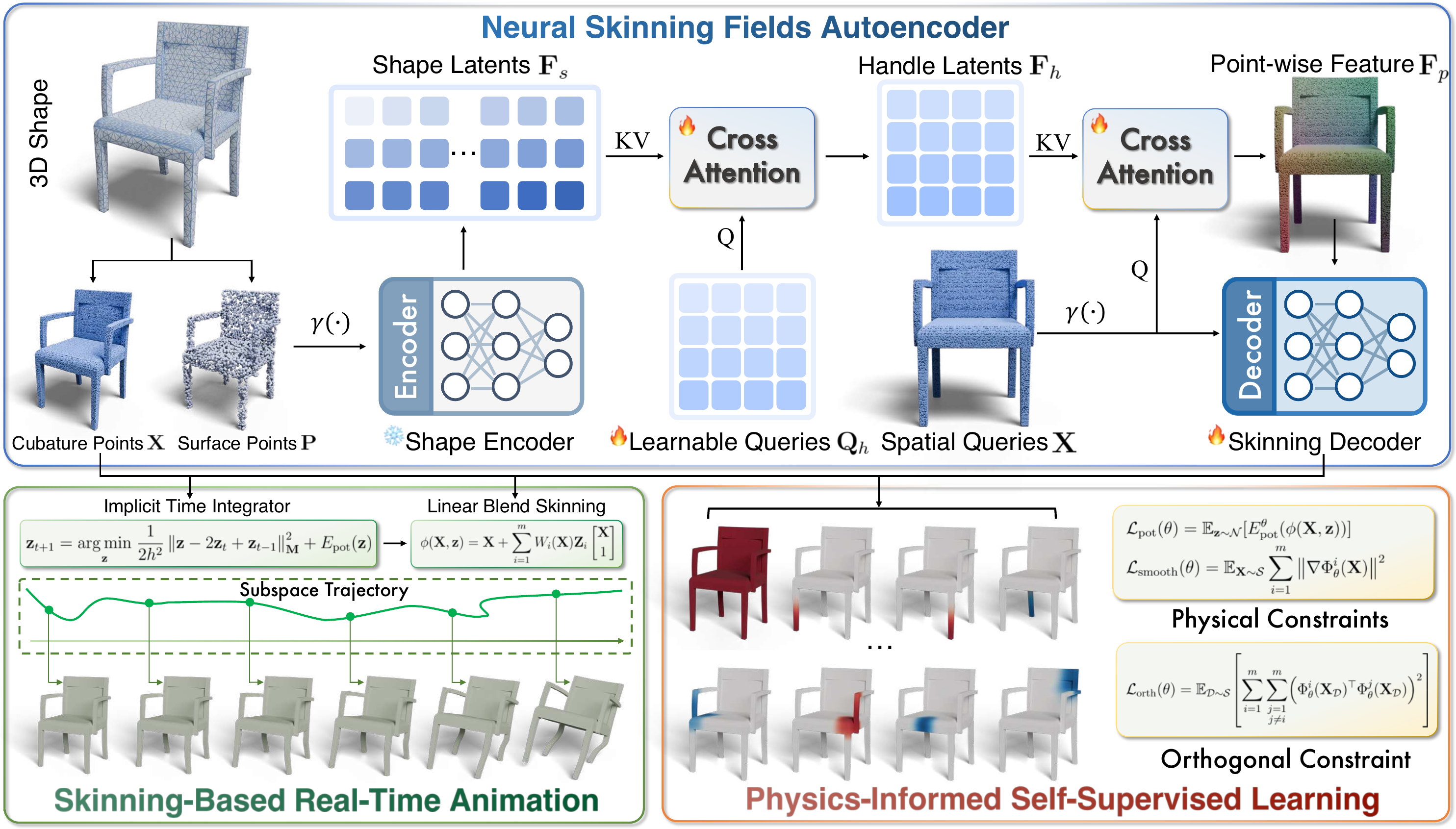}
    \caption{Given a static 3D shape, we first sample volumetric cubature points for animation and surface points for shape encoding. A shape encoder processes the surface points to produce shape latents, from which a set of learnable queries extract handle latents via cross-attention. Subsequently, spatial queries attend to the handle latents through another cross-attention block to derive point-wise features, which are decoded into neural skinning fields. A geometry-only, physics-informed learning strategy optimizes the network in a self-supervised manner under physical and orthogonality constraints, enabling the learned skinning fields to support real-time, physics-based animation.}
    \label{fig:pipeline}
\end{figure*}

\noindent PhysSkin++ is a generalizable physics-informed framework for real-time physics-based animation across diverse 3D shapes and discretizations. As illustrated in Fig.~\ref{fig:pipeline}, our neural skinning fields autoencoder predicts continuous skinning fields from static geometry to define a deformation subspace mapping. The autoencoder is trained through Physics-Informed Self-Supervised Learning (PISSL) under physical and numerical constraints. At runtime, we advance reduced coordinates to animate objects with boundary conditions, interactive control, and contact. Beyond purely physics-driven animation, PhysSkin++ also supports rig-complementary animation by learning complementary deformation subspaces for prescribed rigs. Moreover, the continuous deformation map supports binding to mesh vertices, sampled points, and 3D Gaussians.

Classical full-space physics-based animation methods~\cite{sifakis2012fem, liu2013fast, jiang2016material, xie2024physgaussian, lei2026diffwind, yang2026megas} typically use explicit or implicit time integrators to update the object's motion state. Explicit time integrators are often constrained by stability conditions~\cite{li2025physics}, while implicit time integrators require solving a large nonlinear optimization problem in the high-dimensional full space. PhysSkin++ instead represents the deformation using continuous neural skinning fields and advances only the corresponding reduced-order subspace coordinates through implicit time integration. This lower-dimensional optimization reduces computational cost~\cite{benchekroun2023fast}, enabling real-time physics-based animation.

\noindent\textbf{Continuous Skinning-Based Deformation Representation.}
To model full-space 3D object deformations within a reduced-order subspace, in the spirit of Linear Blend Skinning (LBS)~\cite{magnenat1989joint}, we represent the deformation map $\defX=\DeformationMap(\refX,\dofs(t))$ using a weighted sum of $m$ affine displacement transformations:
\begin{equation}
\DeformationMap(\refX,\dofs)
=\refX+\sum_{i=1}^{\NumHandle}
\W_{\!i}(\refX)\dofsMat_i
\begin{bmatrix}\refX\\1\end{bmatrix}.
\label{eq:unified_deformation_representation}
\end{equation}
Here, $\refX\in\R^3$ is a point in undeformed space and $\defX\in\R^3$ is its deformed position, $m$ is the number of \emph{skinning handles}, $\W_{\!i}:\R^3\rightarrow\R$ is the continuous skinning weight function associated with handle $i$, and $\dofsMat_i\in\R^{3\times4}$ parameterizes its affine displacement. The vector $\dofs=\text{flat}(\dofsMat)\in\R^{12m}$ stacks all handle parameters and serves as the reduced-order subspace coordinates, compactly representing the full-space deformation of $n$ spatial sample points, where $m\ll n$. We first describe the representation and learning procedure for purely physics-driven animation, and then introduce the prescribed rig displacement and rig-space complementarity objective in Sec.~\ref{subsec:rig_complementary_animation}.

\subsection{Neural Skinning Fields Autoencoder}
\label{subsec:skin_autoencoder}
\noindent We train a neural skinning fields autoencoder to predict spatially continuous skinning weights across diverse 3D shapes and discretizations. The resulting mesh-free representation supports the skinning-based real-time animation described in Sec.~\ref{subsec:skinning_based_animation}. Unlike existing neural subspace methods~\cite{modi2024simplicits, chen2023crom} that train a separate neural network for each object, our single feed-forward network generalizes across diverse objects, improving scalability and practicality.

\noindent\textbf{Neural 3D Shape Encoding.}
We use the transformer-based point cloud encoder Michelangelo~\cite{zhao2023michelangelo} to extract a highly compressed latent set $\mathbf{F}_s \in \mathbb{R}^{M \times d}$ for a given 3D shape via a cross-attention block and several self-attention blocks:
\begin{equation}
\mathbf{F}_s = \mathrm{SelfAttn}^{(1:L)}(\mathrm{CrossAttn}(\mathbf{Q}_s, \gamma(\mathbf{P}))),
\label{eq:3dshape_encoder}
\end{equation}
where $\mathbf{Q}_s \in \mathbb{R}^{M \times d}$ is a set of learnable shape latent tokens, $\mathbf{P} \in \mathbb{R}^{N \times 6}$ is the sampled surface point cloud with $N$ points, each point consisting of 3D coordinates and normals, $\gamma(\cdot)$ is point-wise positional encoding, $\mathrm{SelfAttn}^{(1:L)}(\cdot)$ indicates a stack of $L$ self-attention layers that iteratively improve the shape latent representation, $M$ is the number of latents, and $d$ is the dimension of each latent.
In practice, the shape encoder is pretrained on the ShapeNet dataset~\cite{chang2015shapenet} via a signed distance field (SDF) reconstruction task~\cite{park2019deepsdf} to learn generalizable geometric priors. We then freeze the encoder parameters during training to enhance efficiency.

\noindent\textbf{Neural Skinning Fields Decoding.}
To obtain spatially continuous skinning fields, we decode the shape latent set $\mathbf{F}_s$ obtained from Eq.~(\ref{eq:3dshape_encoder}) into spatially varying skinning weights.
We first assign $m$ learnable skinning handle tokens $\mathbf{Q}_h \in \mathbb{R}^{m \times d}$, and then use a cross-attention block to extract a skinning handle latent set $\mathbf{F}_h \in \mathbb{R}^{m \times d}$ from the shape latent set $\mathbf{F}_s$:
\vspace*{0.3em}
\begin{equation}
\mathbf{F}_h = \mathrm{CrossAttn}(\mathbf{Q}_h, \mathrm{SelfAttn}(\mathbf{F}_s)).
\label{eq:skinning_handle_feature}
\vspace{0.3em}
\end{equation}
Next, for any spatial query point $\refX$, we use another cross-attention block to extract a point-wise skinning feature $\mathbf{F}_p \in \mathbb{R}^{d}$ from the skinning handle latent set $\mathbf{F}_h$:
\vspace*{0.3em}
\begin{equation}
\mathbf{F}_p = \mathrm{CrossAttn}(\gamma(\refX), \mathbf{F}_h).
\label{eq:pointwise_skinning_feature}
\vspace{0.3em}
\end{equation}

Finally, we concatenate the point-wise skinning feature $\mathbf{F}_p$ with the positional encoding of the query point, and use a ResNet-style MLP to compute the skinning fields:
\vspace*{0.2em}
\begin{equation}
\W(\refX) = \mathrm{MLP}(\mathbf{F}_p \oplus \gamma(\refX)),
\label{eq:skinning_weights_decoder}
\vspace{0.2em}
\end{equation}
where $\W(\refX) \in \mathbb{R}^{m}$ is the skinning weight vector at point $\refX$, and $\oplus$ denotes the concatenation operator. To promote orthogonality among different skinning modes, we apply an Orthogonalization by Newton's Iteration (ONI) module~\cite{huang2020controllable} at the final layer of the MLP, and we adopt ELU as the activation function in the MLP, which allows learning flexible and expressive skinning-based deformation subspaces without enforcing non-negativity on the learned skinning modes. We denote the full autoencoder by $\Phi_\theta$ and its skinning weight output at $\refX$ by $\Phi_\theta(\refX)$. The continuous fields can be queried throughout the object, while cubature points specify locations for physics-informed training and numerical integration.

\noindent\textbf{Cubature Points Sampling.}
In the classical Finite Element Method (FEM)~\cite{sifakis2012fem}, the object's volume is typically discretized into a tetrahedral mesh~\cite{hu2020fast} to facilitate numerical solutions of physical equations. Unlike triangle surface meshes commonly used for rendering, which only contain surface points, tetrahedral meshes include interior volumetric elements that better capture the object's spatial structure. To train the neural skinning fields autoencoder, we sample cubature points both on the object's surface and inside its volume, with each cubature point serving as a spatial query point $\refX$ for computing the potential energy $\ElasticPotential$ during subspace learning and skinning-based real-time animation. This sampling captures volumetric deformation behavior that surface-only sampling cannot, in contrast to popular 3D Shape VAEs~\cite{zhang20233dshape2vecset} that only sample surface points. To sample interior cubature points, we first convert the object's surface mesh into a watertight mesh~\cite{huang2018robust, huang2020manifoldplus}, then uniformly sample points in the voxelized spatial grid using ray tracing~\cite{3dgrt2024}, and classify points as being outside the object if the ray intersects the surface mesh and the dot product with the normal is negative. For surface cubature points, we employ Sharp Edge Sampling (SES)~\cite{chen2025dora} to better capture geometric details.

\subsection{Physics-Informed Self-Supervised Learning}
\label{subsec:physics_informed_optimization}
\noindent To learn a deformation subspace mapping directly from static geometry without annotated skinning data or precomputed simulation trajectories, we train the neural skinning fields autoencoder with a Physics-Informed Self-Supervised Learning (PISSL) strategy. The learned skinning fields should be spatially smooth and mutually independent, while inducing low-energy deformations. We therefore formulate a self-supervised objective consisting of potential-energy, smoothness, and orthogonality terms:
\begin{equation}
\mathcal{L}_{\mathrm{subspace}}(\theta)
=\mathcal{L}_{\text{smooth}}(\theta)
+\lambda_{\text{pot}}\mathcal{L}_{\text{pot}}(\theta)
+\lambda_{\text{orth}}\mathcal{L}_{\text{orth}}(\theta),
\label{eq:overall_loss}
\end{equation}
where $\lambda_{\text{pot}}$ and $\lambda_{\text{orth}}$ weight the potential-energy and orthogonality losses, respectively. The additional objective for rig-complementary learning is introduced in Sec.~\ref{subsec:rig_complementary_animation}.

\begin{figure}[t]
    \centering
    \includegraphics[width=1.00\linewidth]{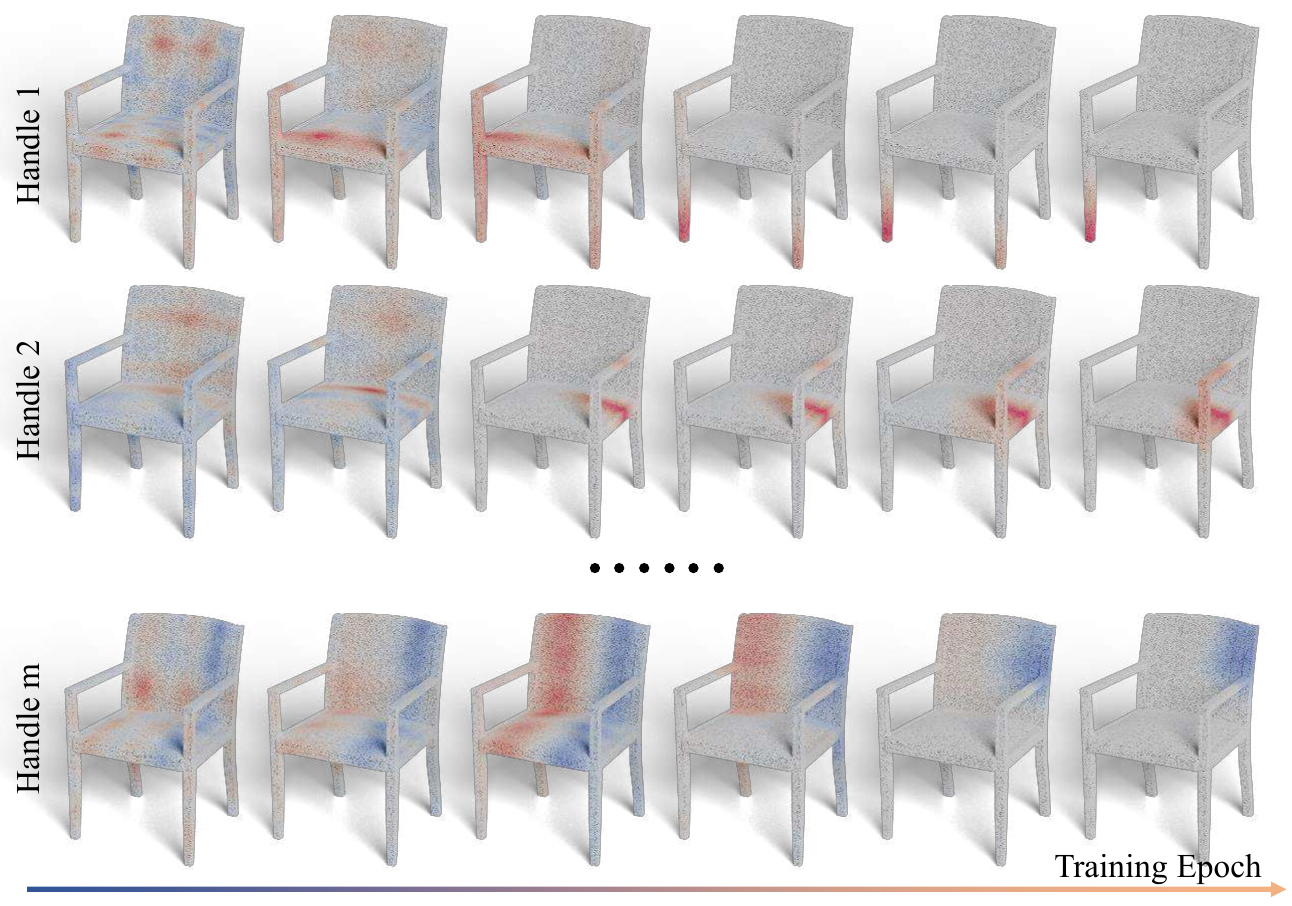}
    \caption{Evolution of neural skinning fields during optimization. Starting from disordered initial representations, our physics-informed self-supervised learning progressively organizes them into physically consistent, approximately orthogonal, and spatially smooth skinning results. For visualization, each skinning weight function $\W_i$ is divided by $\max|\W_i|$ to fall within $[-1, 1]$.}
    \label{fig:ours_training_progress}
\end{figure}

\noindent\textbf{Physical Constraints.}
To encourage physically plausible deformations, we construct a low-energy subspace~\cite{sharp2023data} by minimizing the expected potential energy of deformations induced by randomly sampled subspace coordinates $\dofs \in \R^{12m}$:
\vspace*{0.4em}
\begin{equation}
\mathcal{L}_{\text{pot}}(\theta) = \mathbb{E}_{\dofs \sim \mathcal{N}}[\ElasticPotential^{\theta}(\DeformationMap(\refX, \dofs))],
\label{eq:energy_loss}
\vspace{0.4em}
\end{equation}
where $\dofs$ is defined in Sec.~\ref{subsec:skinning_eigenmodes}, and $\mathcal{N}=\mathcal{N}(\mathbf{0},\mu^2\mathbf{I}_{12m})$ denotes a Gaussian distribution over $\mathbb{R}^{12m}$ with $\mu=0.2$. The expectation is estimated by averaging the potential energy over sampled subspace coordinates within a training batch. Here, $\ElasticPotential^{\theta}(\DeformationMap(\refX,\dofs))$ denotes the object's potential energy evaluated using the deformation map induced by $\dofs$ under the current network parameters $\theta$, with spatial integration approximated using cubature points. During training, we linearly interpolate the potential energy computation from the linear elasticity~\cite{teran2005robust} model to the Neo-Hookean~\cite{kim2020dynamic} model to enhance training stability~\cite{modi2024simplicits}. Besides the low-energy constraint imposed by $\mathcal{L}_{\text{pot}}$, we also expect $\Phi_\theta(\refX)$ to be spatially smooth, so we introduce a spatial smoothness loss:
\begin{equation}
\mathcal{L}_{\text{smooth}}(\theta) = \mathbb{E}_{\refX \sim \mathcal{S}} \sum_{i=1}^{m} \|\nabla \Phi_{\theta}^{i}(\refX)\|^2,
\label{eq:smoothness_loss}
\end{equation}
where $\mathcal{S}$ denotes the set of sampled cubature points introduced in Sec.~\ref{subsec:skin_autoencoder}, and $\nabla \Phi_{\theta}^{i}$ is the spatial gradient of the $i$-th skinning weight function at point $\refX$ under the current network parameters $\theta$.

\noindent\textbf{Orthogonality Constraint.}
To enhance subspace expressiveness, we introduce a loss that promotes orthogonality among the learned skinning weight functions:
\par
{\small
\begin{equation}
\mathcal{L}_{\mathrm{orth}}(\theta)
= \mathbb{E}_{\mathcal{D}\sim\mathcal{S}}
\left[
\sum_{i=1}^{m} \sum_{\substack{j=1\\ j\neq i}}^{m}
\Big( \Phi_{\theta}^{i}(\mathbf{X}_{\mathcal{D}})^\top \Phi_{\theta}^{j}(\mathbf{X}_{\mathcal{D}}) \Big)^2
\right],
\label{eq:orth_sigma_complete}
\end{equation}
}
where $\mathcal{D}$ denotes a training batch of cubature points sampled from the overall set $\mathcal{S}$, and $\mathbf{X}_{\mathcal{D}}$ represents the corresponding point matrix within the batch.

\noindent\textbf{Stable Joint Optimization.}
Simplicits~\cite{modi2024simplicits} jointly penalizes deviations from orthogonality and unit magnitude. We instead $\ell_2$-normalize each column of the skinning weight matrix evaluated at the batch points on the fly. This normalization regulates the scale of the learned modes and prevents numerical drift during training, thereby facilitating optimization of the orthogonality objective and promoting more independent neural skinning fields. However, gradients of the self-supervised loss terms can differ substantially in magnitude and conflict in direction, making naive joint optimization unstable. To overcome this challenge, we complement our on-the-fly skinning-field normalization with ConFIG~\cite{liu2025config} to correct conflicting gradients and stabilize joint optimization.
In Fig.~\ref{fig:ours_training_progress}, we visualize the evolution of different skinning modes during optimization, demonstrating that our physics-informed self-supervised framework progressively organizes the initially unstructured skinning representations into low-energy, approximately orthogonal, and spatially smooth skinning results, which are then used for real-time physics-based animation as detailed in Sec.~\ref{subsec:skinning_based_animation}.

\flushbottom
\subsection{Skinning-Based Real-Time Animation}
\label{subsec:skinning_based_animation}
\label{subsec:controllable_collision_animation}
\noindent The learned neural skinning fields define a continuous subspace mapping through which reduced coordinates determine full-space deformations for physics-based animation. We first describe the basic reduced dynamics, and then incorporate boundary conditions, external control, and contact energies. Rig-complementary dynamics is introduced separately in Sec.~\ref{subsec:rig_complementary_animation}.

\noindent\textbf{Subspace Dynamics.}
With the skinning-based subspace representation, we can create motions with physically plausible deformations directly in the reduced-order subspace. We discretize the governing Newtonian physical equations using standard implicit time integration and update the subspace coordinates $\dofs$ over time as:
\par
{\small
\setlength{\abovedisplayskip}{3pt}
\setlength{\abovedisplayshortskip}{3pt}
\setlength{\belowdisplayskip}{10pt}
\setlength{\belowdisplayshortskip}{10pt}
\begin{equation}
    \dofs_{t+1} =
    \underset{\dofs}{\arg\min}\;
    \frac{1}{2h^2}\|\dofs - 2\dofs_t + \dofs_{t-1}\|_{\mass}^2
    + \ElasticPotential(\DeformationMap(\refX, \dofs)),
    \label{eq:time_stepping}
\end{equation}}
where $h$ is the time step size, $\|\cdot\|_{\mass}$ is the norm induced by the reduced mass matrix $\mass$, and $\ElasticPotential$ is the potential energy that induces the internal force, with $\ElasticPotential$ supporting general hyperelastic material models. This formulation can be solved using standard Newton-based methods~\cite{nocedal2006numerical}. The much lower subspace dimensionality allows efficient optimization of Eq.~(\ref{eq:time_stepping})~\cite{benchekroun2023fast}, enabling real-time physics-based animation.

\begin{figure*}[t]
    \centering
    \includegraphics[width=0.95\textwidth]{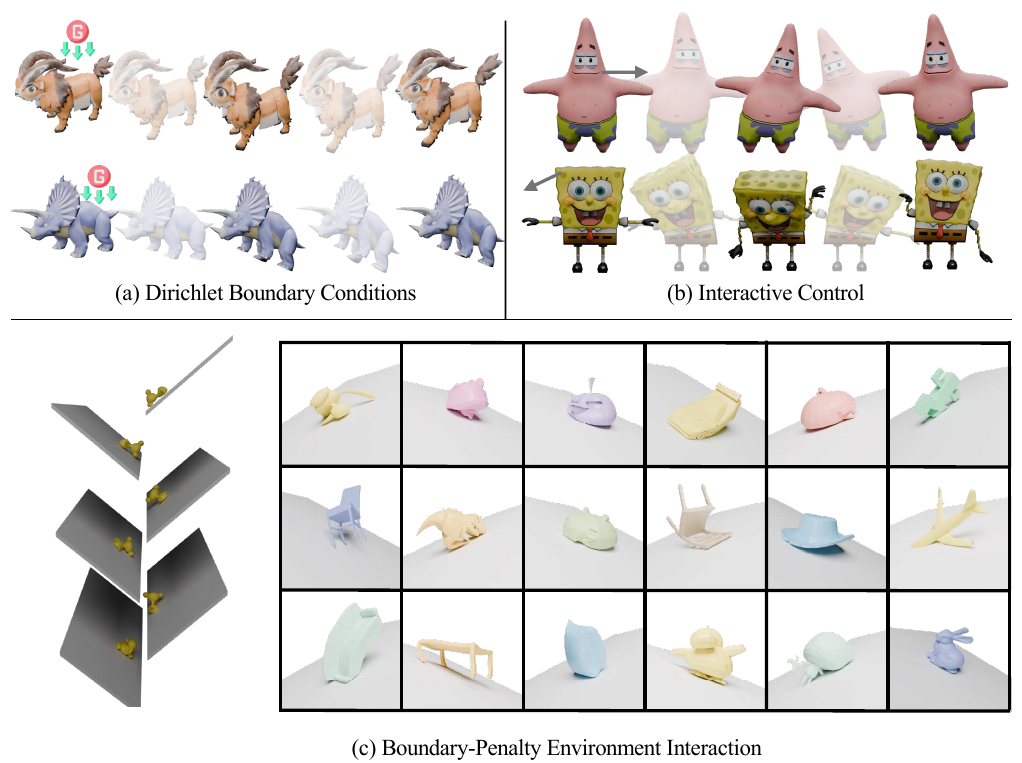}
    \caption{Qualitative results for controllable and collision-aware reduced dynamics. (a) Dirichlet boundary conditions constrain prescribed regions during deformation. (b) Interactive dragging provides localized user control. (c) Boundary-penalty energies support contact with planar environments across diverse shapes. \textbf{We encourage readers to visit our project page for additional animation videos.}}
    \label{fig:controllable_reduced_animation}
    \vspace{-0.8em}
\end{figure*}

\noindent\textbf{Reduced Dynamics with Control and Contact.}
We extend Eq.~(\ref{eq:time_stepping}) by augmenting its elastic potential with the energies associated with each animation setting. Let $\DirichletMap$ denote the deformation map satisfying the prescribed Dirichlet boundary conditions. For a fixed boundary displacement field and fixed constrained region, we update the reduced coordinates via:
\begin{equation}
\begin{aligned}
\dofs_{t+1}=\underset{\dofs}{\arg\min}\;&
\frac{1}{2h^2}
\left\|\dofs-2\dofs_t+\dofs_{t-1}\right\|_{\mass}^{2}
+\ElasticPotential\!\left(\DirichletMap(\refX,\dofs)\right)\\
&+\ForceEnergy(\dofs,t)
+\BoundaryEnergy(\dofs)
+\CollisionEnergy(\dofs).
\end{aligned}
\label{eq:unified_reduced_dynamics}
\end{equation}
Here, $\ForceEnergy$ accounts for the effects of gravity or user-defined actuation forces, $\BoundaryEnergy$ models penalty-based environment interaction, and $\CollisionEnergy$ accounts for inter-object contact. In the absence of boundary constraints, $\DirichletMap$ reduces to $\DeformationMap$. Activating the relevant terms in Eq.~(\ref{eq:unified_reduced_dynamics}) supports controllable and collision-aware animation.

\noindent\textbf{Boundary and External Control.}
We enforce prescribed boundary displacements through Dirichlet boundary conditions and project global or localized external forces into the reduced space to drive and interactively control object deformation.

\noindent\textit{Dirichlet Boundary Conditions.}
Let $\Domain\subset\R^3$ denote the object's rest domain and $\DirichletBoundary\subseteq\partial\Domain$ its Dirichlet boundary. We define an indicator $d:\Domain\rightarrow\{0,1\}$ for the unconstrained region, with $d=0$ on $\DirichletBoundary$ and $d=1$ elsewhere, and let $\BoundaryValue:\Domain\rightarrow\R^3$ be a displacement field that matches the prescribed displacement on $\DirichletBoundary$. Following a boundary-compatible construction, we modify the deformation map in Eq.~(\ref{eq:unified_deformation_representation}) as:
\begin{equation}
\DirichletMap(\refX,\dofs)
=\refX+\BoundaryValue(\refX)
+d(\refX)\sum_{i=1}^{\NumHandle}
\W_{\!i}(\refX)\dofsMat_i
\begin{bmatrix}\refX\\1\end{bmatrix}.
\label{eq:dirichlet_deformation_map}
\end{equation}
This construction is equivalent to modulating the runtime skinning fields by $d(\refX)$. Consequently, the displacement induced by the reduced coordinates vanishes on $\DirichletBoundary$, where $\BoundaryValue$ directly specifies the prescribed boundary displacement for arbitrary reduced coordinates. Different boundary configurations are specified by selecting different subsets of $\partial\Domain$. As shown in Fig.~\ref{fig:controllable_reduced_animation}(a), varying the constrained region produces distinct boundary-conditioned deformation behavior under the same reduced formulation.

\noindent\textit{External Force and Interactive Control.}
External controls are represented as full-space force fields and then mapped to the reduced coordinates through the deformation map. Let $\Jacobian_\phi$ denote the Jacobian of the active deformation map with respect to the reduced coordinates. A full-space force $\mathbf{f}_x(t)$ gives:
\begin{equation}
\mathbf{f}_z(t)=\Jacobian_\phi^{\top}\mathbf{f}_x(t),
\qquad
\ForceEnergy(\dofs,t)=-\mathbf{f}_z(t)^{\top}\dofs.
\label{eq:force_projection}
\end{equation}
This linear potential applies to forces held fixed during each reduced solve. It covers global force fields, such as gravity, and localized user-defined actuation forces. Different controls share the same force-to-coordinate mapping while differing in how $\mathbf{f}_x(t)$ is constructed. For interactive dragging, we construct a Gaussian-kernel-weighted local force around the selected point and update the drag target position from the current user input at each frame before applying Eq.~(\ref{eq:force_projection}). As shown in Fig.~\ref{fig:controllable_reduced_animation}(b), localized dragging enables direct manipulation of different character geometries.

\noindent\textbf{Contact-Aware Multi-Object Dynamics.}
We handle contact through boundary-penalty environment interaction and point-based IPC-style collision energies, with the latter supporting both rigid--deformable and deformable--deformable interaction.

\noindent\textit{Boundary-Penalty Environment Interaction.}
We model simple environmental interactions, such as plane collision, as penalty constraints to prescribed geometric boundaries. Let $\Gamma_E$ denote an environment boundary, and let $\mathcal{A}$ be the active set of animation points within its influence region. For each active point $j$, we compute a boundary target $\bar{\mathbf{x}}_j\in\Gamma_E$ from the current environment geometry. The boundary penalty is:
\begin{equation}
\BoundaryEnergy(\dofs)
=\sum_{j\in\mathcal{A}}
\left\|
\DirichletMap(\refX_j,\dofs)
-\bar{\mathbf{x}}_j
\right\|_2^2.
\label{eq:boundary_penalty_energy}
\end{equation}
The active set and targets are held fixed when differentiating this energy. The resulting full-space derivatives are mapped to the reduced coordinates using the same deformation-map Jacobian $\Jacobian_\phi$ as in Eq.~(\ref{eq:force_projection}). For plane-collision scenarios, such as an inclined ramp, $\Gamma_E$ is instantiated as a finite rectangular boundary with a prescribed normal. The target $\bar{\mathbf{x}}_j$ is obtained by projecting the current point onto the boundary plane and restricting it to the finite boundary extent. Moving or rotating the boundary updates its geometry, active points, and boundary targets. As shown in Fig.~\ref{fig:controllable_reduced_animation}(c), the same boundary-penalty formulation supports contact with planar environments across diverse object geometries and boundary orientations.

\begin{figure*}[t]
    \centering
    \includegraphics[width=0.95\textwidth]{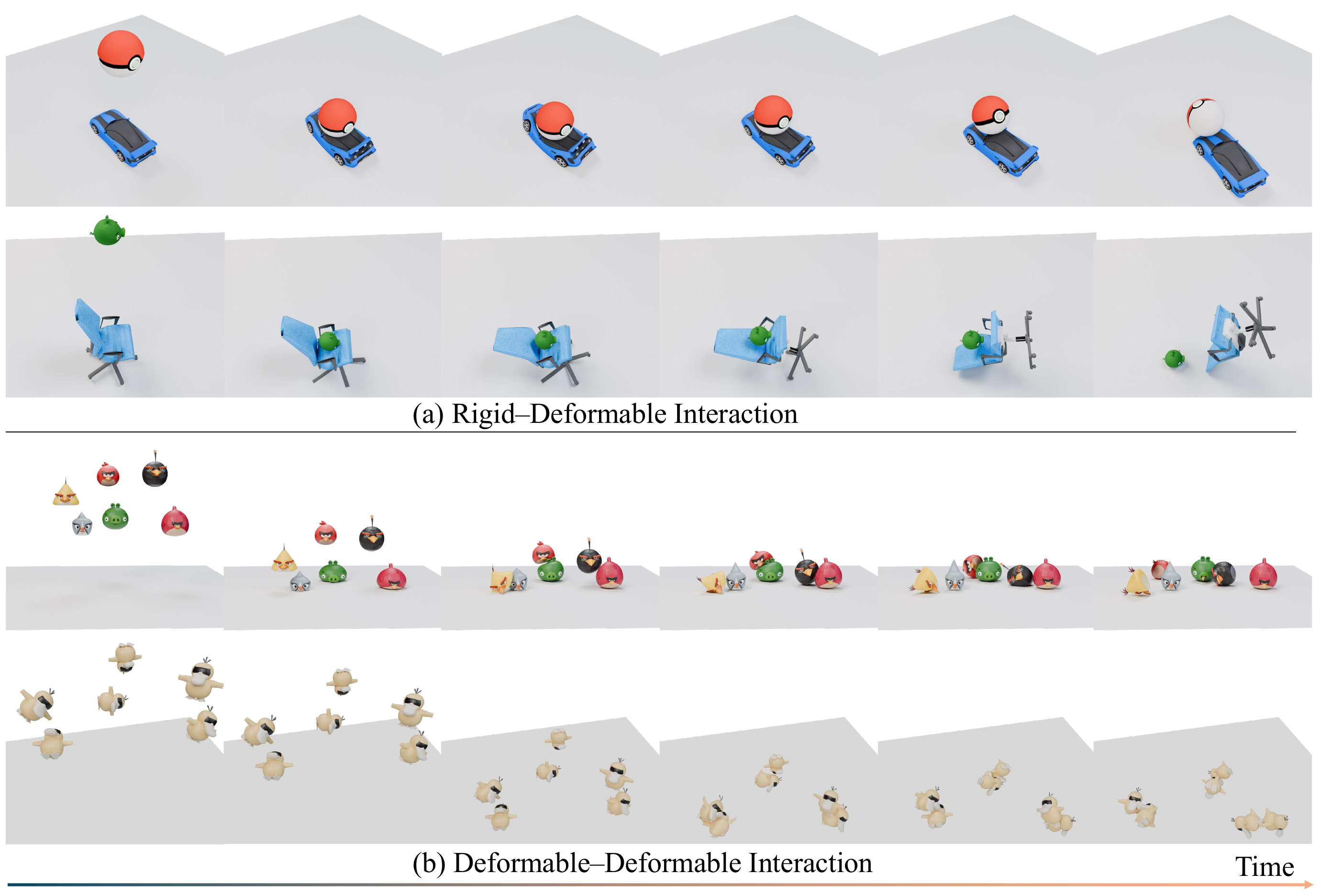}
    \caption{Qualitative results for collision-aware multi-object animation. (a) Rigid--deformable contact between objects with distinct geometries. (b) Deformable--deformable contact among multiple objects, jointly resolved by the reduced collision formulation. \textbf{We encourage readers to visit our project page for additional animation videos.}}
    \label{fig:multi_object_interaction}
    \vspace{-0.8em}
\end{figure*}

\noindent\textit{IPC-Style Collision Energy.}
For multi-object interaction, we use sampled integration points as collision primitives and construct a contact set $\mathcal{C}$ by querying neighboring points from different objects. For each contact $c$, we store a contact direction $\mathbf{n}_c$ and define the contact gap $d_c$ as the signed separation along this direction. We normalize this gap by the contact radius $r_c>0$, i.e., $\widehat d_c=d_c/r_c$, and use the following point-based log barrier inspired by Incremental Potential Contact (IPC)~\cite{li2020ipc}:
\begin{equation}
E_{b,c}=
\begin{cases}
-(\widehat d_c-1)^2\log(\widehat d_c-r_p),
&r_p<\widehat d_c<1,\\
0,&\widehat d_c\geq1,\\
+\infty,&\widehat d_c\leq r_p.
\end{cases}
\label{eq:point_ipc_barrier}
\end{equation}
We assemble the collision term as $\CollisionEnergy=k_c\sum_{c\in\mathcal{C}}E_{b,c}$, defining an IPC-style barrier over sampled point pairs. Here, $0\leq r_p<1$ specifies the minimum normalized contact gap, and $k_c$ controls the contact stiffness. The barrier is finite only for $\widehat d_c>r_p$.

\noindent\textit{Rigid--Deformable Interaction.}
We model rigid-body motion with a global rigid transformation, while the deformable geometry evolves according to the neural skinning deformation map. At the beginning of each time step, the transformed rigid surface samples and current deformable points are used to construct the active contact pairs. For each pair, the point-pair barrier in Eq.~(\ref{eq:point_ipc_barrier}) is evaluated from the contact gap between the rigid and deformable sides. The rigid transformation is treated as prescribed input to this reduced deformable solve. As shown in Fig.~\ref{fig:multi_object_interaction}(a), this formulation handles rigid--deformable contact across different object pairs and motions.

\noindent\textit{Deformable--Deformable Interaction.}
For $K$ deformable objects, each object $i$ maintains its own reduced coordinates $\dofs_i$, material parameters, and subspace dimension. We concatenate the reduced coordinates into the global reduced variable $\mathbf{z}=[\dofs_1^{\top},\ldots,\dofs_K^{\top}]^{\top}$ and sum the per-object inertial, elastic, and control energies. Cross-object contacts couple these states through the same point-pair barrier in Eq.~(\ref{eq:point_ipc_barrier}). For a contact $c$ between objects $a$ and $b$, let $\Jacobian_{\phi,a}$ and $\Jacobian_{\phi,b}$ denote their deformation-map Jacobians with respect to $\dofs_a$ and $\dofs_b$, respectively, evaluated at the corresponding contact points. With the contact direction and radius held fixed during differentiation, the contact gap is affine in $\mathbf{z}$. Applying the chain rule to the barrier energy gives the following reduced derivatives, showing only the nonzero blocks for objects $a$ and $b$:
\begin{equation}
\begin{aligned}
\Jacobian_c&=\frac{\partial d_c}{\partial\mathbf{z}}
=\mathbf{n}_c^{\top}[\,\Jacobian_{\phi,a}\;-\Jacobian_{\phi,b}\,],\\
\nabla_{\mathbf{z}}E_{b,c}
&=\Jacobian_c^{\top}\frac{\partial E_{b,c}}{\partial d_c},
\qquad
\nabla_{\mathbf{z}}^2E_{b,c}
=\Jacobian_c^{\top}
\frac{\partial^2 E_{b,c}}{\partial d_c^2}
\Jacobian_c.
\end{aligned}
\label{eq:multi_object_contact_derivatives}
\end{equation}
At each reduced solve step, we assemble these per-contact derivatives into the global gradient and Hessian of $\CollisionEnergy(\mathbf{z})$, directly coupling the reduced states of contacting objects. As shown in Fig.~\ref{fig:multi_object_interaction}(b), the resulting coupled system jointly resolves contacts among multiple deformable objects.

\subsection{Rig-Complementary Animation}
\label{subsec:rig_complementary_animation}
\noindent Beyond purely physics-driven animation, PhysSkin++ also supports rig-complementary animation when an artist-authored LBS rig is available. The prescribed rig deformation serves as the primary animation, while complementary elastic deformation is parameterized by reduced coordinates through the learned neural skinning fields. We first describe the rig-based deformation and then introduce the rig-space complementarity objective used during self-supervised learning.

\noindent\textbf{Rig-Based Deformation Representation.}
Consider a skeletal rig with $n_{\mathrm b}$ bones, pose parameters $\mathbf{p}$, and rest pose $\mathbf{p}^0$. Let $\mathbf{G}_b(\mathbf{p})\in\mathrm{SE}(3)$ be the global transformation of bone $b$, obtained by composing its local transformations along the kinematic hierarchy. We denote by $\mathbf{A}_b(\mathbf{p})\in\mathbb{R}^{3\times4}$ the affine transform induced by the relative bone motion $\mathbf{G}_b(\mathbf{p})\mathbf{G}_b(\mathbf{p}^0)^{-1}$. With rig skinning weights $\W_b^{\mathrm{rig}}(\refX)$, LBS~\cite{magnenat1989joint} maps a rest-space point as:
\begin{equation}
\DeformationMap_{\mathrm{rig}}(\refX,\mathbf{p})
=\sum_{b=1}^{n_{\mathrm b}}
\W_b^{\mathrm{rig}}(\refX)
\mathbf{A}_b(\mathbf{p})
\begin{bmatrix}\refX\\1\end{bmatrix}.
\label{eq:rig_lbs_deformation}
\end{equation}
Here, $\mathbf{A}_b(\mathbf{p})$ moves the point from the rest configuration according to bone $b$, and the weighted sum blends the bone-transformed positions. Subtracting the rest position gives the rig displacement $\mathbf{u}_{\mathrm{rig}}(\refX,\mathbf{p})=\DeformationMap_{\mathrm{rig}}(\refX,\mathbf{p})-\refX$. This formulation is efficient and directly controllable, but its kinematic deformations alone cannot produce physical elastic responses.

For rig-complementary animation, we augment the representation in Eq.~(\ref{eq:unified_deformation_representation}) with the prescribed rig displacement:
\begin{equation}
\begin{aligned}
\DeformationMap(\refX,\mathbf{p},\dofs)
=\;&\refX+\mathbf{u}_{\mathrm{rig}}(\refX,\mathbf{p})\\
&+\sum_{i=1}^{\NumHandle}
\W_{\theta,i}^{\mathrm{comp}}(\refX)\dofsMat_i
\begin{bmatrix}\refX\\1\end{bmatrix}.
\end{aligned}
\label{eq:rig_complementary_deformation_representation}
\end{equation}
The neural skinning term defines the complementary elastic displacement $\mathbf{u}_{\mathrm{comp}}(\refX,\dofs)$. Both animation settings share the continuous skinning parameterization, with rig-complementary animation incorporating the prescribed rig displacement.

\begin{figure}[t]
    \centering
    \includegraphics[width=0.98\linewidth]{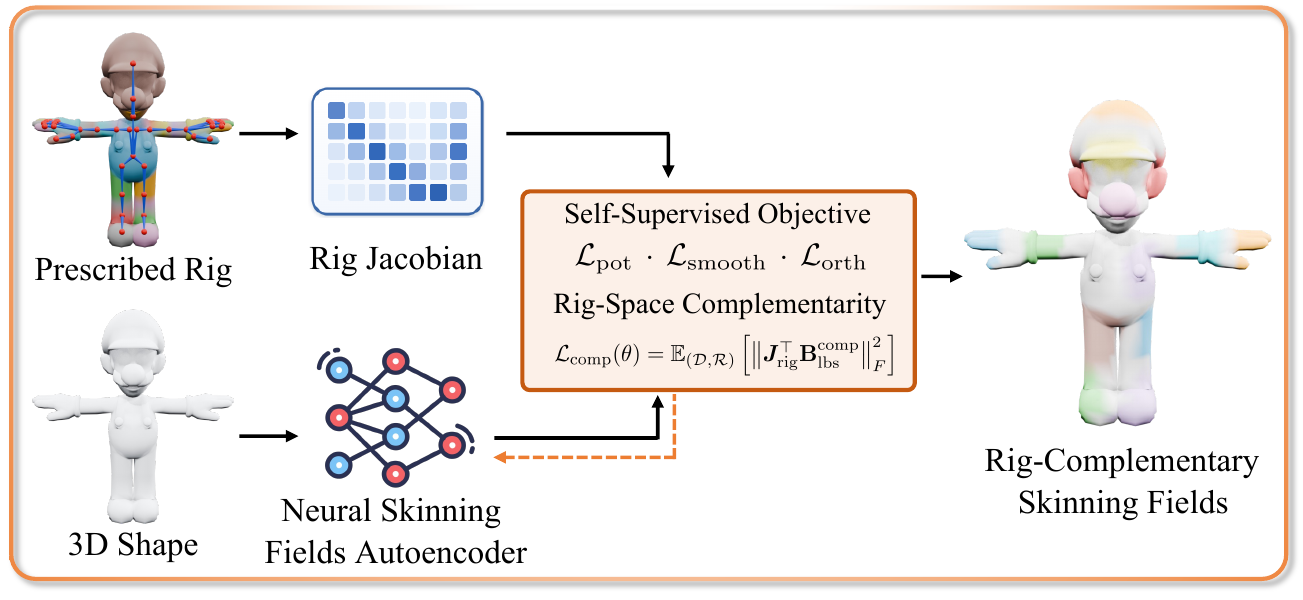}
    \caption{Self-supervised rig-complementarity training. The rig Jacobian of a prescribed rig provides a complementarity constraint for learning rig-complementary neural skinning fields from static geometry.}
    \label{fig:rig_comp_training_pipeline}
\end{figure}

\noindent\textbf{Rig-Space Complementarity Constraint.}
Fig.~\ref{fig:rig_comp_training_pipeline} summarizes our self-supervised rig-complementarity training pipeline.
For a batch $\mathcal{D}$ of $N_{\mathcal{D}}$ cubature points, we stack the rig and complementary displacements as $\mathbf{u}_{\mathrm{rig}}(\mathbf{p}),\mathbf{u}_{\mathrm{comp}}(\dofs)\in\mathbb{R}^{3N_{\mathcal{D}}}$. We denote the rig Jacobian by $\Jacobian_{\mathrm{rig}}=\partial\mathbf{u}_{\mathrm{rig}}(\mathbf{p})/\partial\mathbf{p}$. To prevent the physical component from duplicating motion expressible by the rig, we seek complementary displacements satisfying:
\begin{equation}
\Jacobian_{\mathrm{rig}}^{\top}\mathbf{u}_{\mathrm{comp}}=\mathbf{0}.
\label{eq:rig_space_complementarity}
\end{equation}
This condition makes the complementary displacement orthogonal to the rig motion subspace spanned by the columns of $\Jacobian_{\mathrm{rig}}$ under the Euclidean inner product.

To impose this condition for arbitrary subspace coordinates $\dofs$, we arrange the batch point matrix as $\mathbf{X}_{\mathcal{D}}\in\R^{3\times N_{\mathcal{D}}}$ and collect the predicted complementary weights in $\W_{\theta}^{\mathrm{comp}}(\mathbf{X}_{\mathcal{D}})\in\R^{N_{\mathcal{D}}\times m}$. Consistent with Eq.~(\ref{eq:rig_complementary_deformation_representation}), the corresponding discrete skinning matrix is constructed as:
\par
{\small
\begin{equation}
\begin{aligned}
\mathbf{B}_{\mathrm{lbs}}^{\mathrm{comp}}
&=\mathbf{I}_3\otimes\Bigg[
\left(
\mathbf{1}_{m}^{\top}\otimes
\begin{bmatrix}
\mathbf{X}_{\mathcal{D}}\\
\mathbf{1}_{N_{\mathcal{D}}}^{\top}
\end{bmatrix}^{\top}
\right)\\
&\qquad\odot
\left(
\W_{\theta}^{\mathrm{comp}}(\mathbf{X}_{\mathcal{D}})
\otimes\mathbf{1}_{4}^{\top}
\right)
\Bigg].
\end{aligned}
\label{eq:learned_complementarity}
\end{equation}
}
Here, $\otimes$ and $\odot$ denote the Kronecker and Hadamard products, respectively, and $\mathbf{B}_{\mathrm{lbs}}^{\mathrm{comp}}\in\R^{3N_{\mathcal{D}}\times12m}$. The displacement stacking and the flattening of handle parameters use the ordering consistent with this matrix, so that $\mathbf{u}_{\mathrm{comp}}=\mathbf{B}_{\mathrm{lbs}}^{\mathrm{comp}}\dofs$.

We use a linear control parameterization of the prescribed rig, yielding a constant rig Jacobian $\Jacobian_{\mathrm{rig}}$~\cite{benchekroun2023fast}. Since $\mathbf{u}_{\mathrm{comp}}=\mathbf{B}_{\mathrm{lbs}}^{\mathrm{comp}}\dofs$, the complementarity condition in Eq.~(\ref{eq:rig_space_complementarity}) holds for arbitrary $\dofs$ when:
\begin{equation}
\Jacobian_{\mathrm{rig}}^{\top}
\mathbf{B}_{\mathrm{lbs}}^{\mathrm{comp}}
=\mathbf{0}.
\label{eq:learned_complementarity_constraint}
\end{equation}
We optimize the predicted complementary skinning weights using the following rig-space complementarity loss:
\begin{equation}
\mathcal{L}_{\mathrm{comp}}(\theta)
=\mathbb{E}_{(\mathcal{D},\mathcal{R})}
\left[
\left\|
\Jacobian_{\mathrm{rig}}^{\top}
\mathbf{B}_{\mathrm{lbs}}^{\mathrm{comp}}
\right\|_F^2
\right],
\label{eq:self_supervised_complementarity_loss}
\end{equation}
where $\mathcal{R}$ denotes the prescribed rig and its Jacobian is evaluated on the same batch points as $\mathbf{B}_{\mathrm{lbs}}^{\mathrm{comp}}$. The rig-complementary training objective augments Eq.~(\ref{eq:overall_loss}):
\begin{equation}
\mathcal{L}_{\mathrm{rig}}(\theta)
=\mathcal{L}_{\mathrm{subspace}}(\theta)
+\lambda_{\mathrm{comp}}\mathcal{L}_{\mathrm{comp}}(\theta),
\label{eq:rig_overall_loss}
\end{equation}
where $\mathcal{L}_{\mathrm{subspace}}$ is the self-supervised subspace learning objective defined in Eq.~(\ref{eq:overall_loss}), $\mathcal{L}_{\mathrm{comp}}$ is the rig-space complementarity loss defined in Eq.~(\ref{eq:self_supervised_complementarity_loss}), and $\lambda_{\mathrm{comp}}$ is a weighting hyperparameter. We optimize the resulting objective using the on-the-fly skinning-field normalization and conflict-aware gradient correction described in Sec.~\ref{subsec:physics_informed_optimization}.

\begin{figure*}[t]
    \centering
    \includegraphics[width=0.95\textwidth]{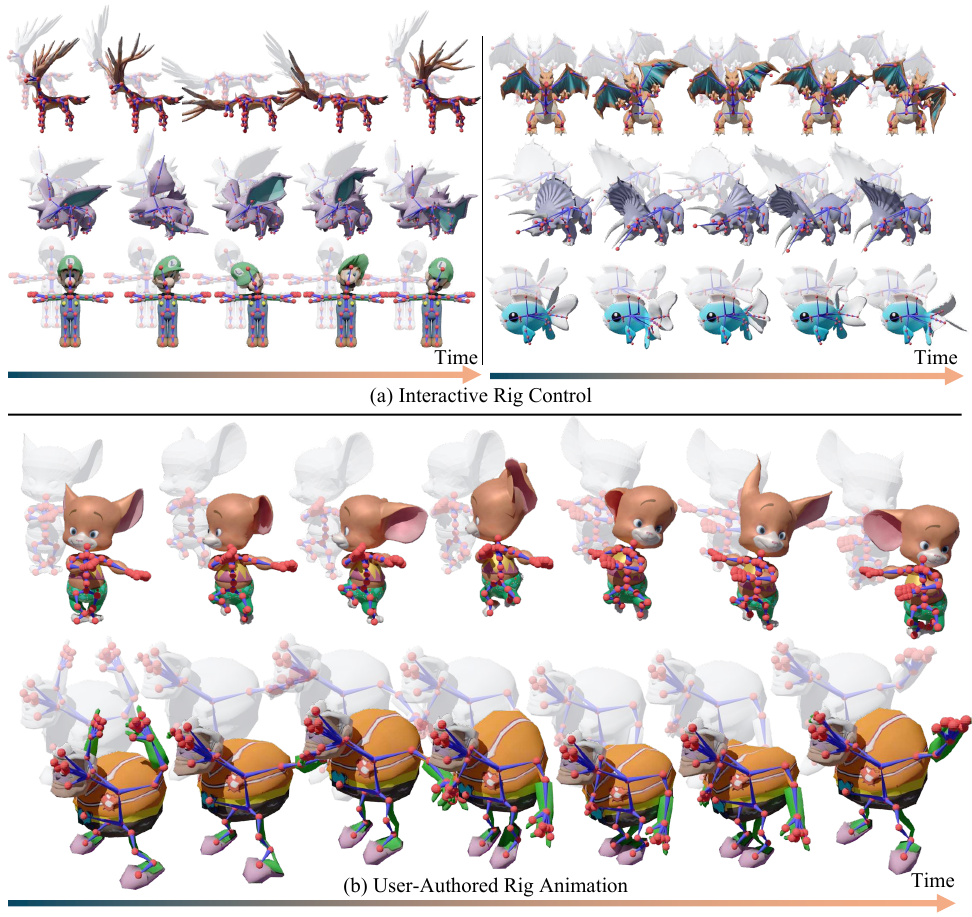}
    \caption{Qualitative results of rig-complementary animation. For each frame, the rig-driven primary motion without complementary elastic deformation is shown semi-transparently as a reference to the corresponding rig-complementary animation. (a) During interactive rig control, user-manipulated poses define the primary animation, and PhysSkin++ produces the corresponding complementary elastic response at runtime. (b) For user-authored rig animation sequences, PhysSkin++ augments the input kinematic trajectory with temporally evolving complementary elastic deformation. \textbf{We encourage readers to visit our project page for additional animation videos.}}
    \label{fig:rig_complementary_animation}
    \vspace{-0.8em}
\end{figure*}

\noindent\textbf{Rig-Driven Complementary Dynamics.}
\label{subsec:rig_complementary_dynamics}
At runtime, the deformation representation in Eq.~(\ref{eq:rig_complementary_deformation_representation}) expresses object deformation through the prescribed rig displacement and a complementary elastic displacement parameterized by the reduced coordinates. The rig trajectory $\{\mathbf{p}_t\}$ is supplied as an external kinematic input, such as an animator-authored pose sequence. Given the prescribed pose $\mathbf{p}_{t+1}$, we update the complementary coordinates through implicit reduced dynamics:
\begin{equation}
\begin{aligned}
\dofs_{t+1}&=\underset{\dofs}{\arg\min}\;
\frac{1}{2h^2}
\left\|\dofs-2\dofs_t+\dofs_{t-1}\right\|_{\mass}^{2}\\
&\quad+\ElasticPotential\!\left(
\refX+\mathbf{u}_{\mathrm{comp}}(\refX,\dofs)
+\mathbf{u}_{\mathrm{rig}}(\refX,\mathbf{p}_{t+1})
\right).
\end{aligned}
\label{eq:rig_complementary_time_stepping}
\end{equation}
The complementarity loss in Eq.~(\ref{eq:self_supervised_complementarity_loss}) encourages the condition in Eq.~(\ref{eq:rig_space_complementarity}) during training, allowing the runtime solve to optimize only the complementary coordinates without additional complementarity constraints. As shown in Fig.~\ref{fig:rig_complementary_animation}, this formulation supports both interactive rig control and user-authored rig animation sequences, enriching their kinematic deformation with physics-based complementary elastic response.

\subsection{Discretization-Agnostic Animation Binding}
\label{subsec:gaussian_binding}
\noindent A key advantage of representing skinning weights as continuous neural fields is the resulting discretization-agnostic deformation map. The learned fields can be evaluated directly at mesh vertices, sampled points, and neural scene primitives, allowing the same reduced physical state to deform different spatial discretizations. We demonstrate this binding capability with 3D Gaussian Splatting by directly binding the anisotropic Gaussian primitives to the learned skinning fields.

\noindent\textbf{3D Gaussian Representation.}
3D Gaussian Splatting (3DGS)~\cite{kerbl20233d} represents a scene using anisotropic Gaussian primitives $\mathcal{G}=\{\mathcal{G}_g\}_{g=1}^{N_G}$, each parameterized as:
\begin{equation}
\mathcal{G}_g=
\left\{
\boldsymbol{\mu}_g,\,
\mathbf{R}_g,\,
\mathbf{s}_g,\,
\alpha_g,\,
\mathbf{c}_g
\right\},
\label{eq:gaussian_primitive}
\end{equation}
where $\boldsymbol{\mu}_g\in\R^3$ is the center, $\mathbf{R}_g\in\mathrm{SO}(3)$ is the rotation matrix, $\mathbf{s}_g\in\R_{+}^3$ contains the principal-axis scales, $\alpha_g\in[0,1]$ is the opacity, and $\mathbf{c}_g$ denotes the spherical harmonic (SH) appearance coefficients. The corresponding 3D covariance matrix is:
\begin{equation}
\Sigma_g
=\mathbf{R}_g\mathbf{S}_g\mathbf{S}_g^{\top}\mathbf{R}_g^{\top},
\qquad
\mathbf{S}_g=\operatorname{diag}(\mathbf{s}_g).
\label{eq:gaussian_covariance_parameterization}
\end{equation}
For rendering, each 3D Gaussian is projected to a screen-space Gaussian, whose color contribution is accumulated through depth-ordered alpha blending. Animating the Gaussian representation therefore requires updating the center and covariance of each Gaussian according to the underlying physical deformation.

\noindent\textbf{Continuous Gaussian Binding.}
Let $\boldsymbol{\mu}_g^0$ denote the rest center of the $g$-th Gaussian. Since the neural skinning fields are spatially continuous, we directly evaluate $\W_i(\boldsymbol{\mu}_g^0)$ at its rest center. Following the deformation map in Eq.~(\ref{eq:unified_deformation_representation}), we construct the local affine transformation for each Gaussian as:
\begin{equation}
\mathbf{T}_g(\dofs)
\equiv
\left[\mathbf{A}_g(\dofs)\;\mathbf{t}_g(\dofs)\right]
=\left[\mathbf{I}_3\;\mathbf{0}\right]
+\sum_{i=1}^{\NumHandle}
\W_{\!i}(\boldsymbol{\mu}_g^0)\dofsMat_i,
\label{eq:gaussian_local_affine_map}
\end{equation}
where $\mathbf{A}_g\in\R^{3\times3}$ is the linear part of the affine transformation, which may include rotation, scaling, and shear, and $\mathbf{t}_g\in\R^3$ is its translation. We evaluate the skinning weights once at each Gaussian's rest center and use these fixed weights throughout its kernel. The resulting local affine transformation is updated from the current reduced coordinates.

\noindent\textbf{Gaussian Kernel Deformation.}
Let $\Sigma_g^0$ denote the rest covariance. Applying the local affine transformation $\mathbf{T}_g(\dofs)$ to the Gaussian kernel gives:
\begin{equation}
\begin{aligned}
\boldsymbol{\mu}_g(\dofs)
&=\mathbf{A}_g(\dofs)\boldsymbol{\mu}_g^0
+\mathbf{t}_g(\dofs),\\
\Sigma_g(\dofs)
&=\mathbf{A}_g(\dofs)\Sigma_g^0
\mathbf{A}_g(\dofs)^{\top}.
\end{aligned}
\label{eq:gaussian_kernel_update}
\end{equation}
The center therefore follows the skinned point motion, while the covariance captures the induced rotation and anisotropic deformation of the Gaussian kernel. The opacity and SH coefficients retain their rest values.

\noindent\textbf{Runtime Coupling.}
At runtime, we advance the reduced coordinates $\dofs$ using Eq.~(\ref{eq:time_stepping}). At each frame, these coordinates determine the local affine maps in Eq.~(\ref{eq:gaussian_local_affine_map}), which update the Gaussian centers and covariances through Eq.~(\ref{eq:gaussian_kernel_update}), driving physics-based animation of the Gaussian representation.

\setlength{\textfloatsep}{12pt plus 2pt minus 2pt}
\setlength{\floatsep}{8pt plus 2pt minus 2pt}
\section{Experiments}
\label{sec:experiments}
\subsection{Implementation Details}
\label{subsec:Implementation}
\noindent For the 3D shape encoder, we adopt the pre-trained Michelangelo~\cite{zhao2023michelangelo} model which consists of one cross-attention block and eight self-attention blocks; the output latent set $\mathbf{F}_s \in \mathbb{R}^{M \times d}$ has $M = 256$ and $d=768$. For each 3D object, we re-center and normalize it into a bounding volume of $[-1, 1]^3$. For the 3D shape encoder, we sample 4096 points on the object surface as input. For cubature points sampling, we first sample 100k points on the surface, and then voxelize the space into a $128^3$ grid to obtain interior points to form the complete candidate cubature point set. For each training batch, we randomly sample 1000 cubature points from the candidate point set for potential energy computation, following the same strategy as previous work~\cite{chang2023licrom, modi2024simplicits}. In each batch, we randomly sample 1024 subspace coordinate vectors $\dofs$ for training. Our framework is trained on 4 NVIDIA GeForce RTX 4090 GPUs. The learning rate is linearly increased to $5\times 10^{-4}$ within the first $1\%$ of training iterations (warm-up), and then gradually decreased using the cosine decay schedule until reaching the minimum value of $5\times10^{-5}$.

\subsection{Evaluation of Learned Skinning Fields}
\noindent\textbf{Baselines.}
We compare our method with the following baselines: Simplicits~\cite{modi2024simplicits}, RigNet~\cite{xu2020rignet}, M-I-A~\cite{guo2025make}, Anymate~\cite{deng2025anymate}, and Puppeteer~\cite{song2025puppeteer} on the ShapeNet~\cite{chang2015shapenet} and RigNet~\cite{xu2020rignet} datasets. Simplicits trains a separate neural network for each object to output skinning fields through self-supervised learning. The other baselines are supervised neural rigging and skinning methods that output object joints, bones, and skinning weights in a feed-forward manner. These methods are pretrained on expert-annotated datasets, then fine-tuned for comparison.

\begin{table}[t]
\centering
\caption{
Quantitative comparison on the RigNet~\cite{xu2020rignet} dataset.
}
\resizebox{0.95\linewidth}{!}{
\tabcolsep 8pt
\footnotesize
\begin{tabular}{lccc}
\toprule
\multicolumn{1}{c}{Method} & \multicolumn{1}{l}{$\Omega_{\text{orth}} \times 10^{-2}$ $\downarrow$} & \multicolumn{1}{l}{$\kappa_{\log}$ $\downarrow$} & \multicolumn{1}{l}{$H_{\text{spec}}$ $\uparrow$}  \\ 
\midrule
RigNet~\cite{xu2020rignet} & \cellcolor[HTML]{FAC791}0.5324 & \cellcolor[HTML]{FFFF99}2.7997  & \cellcolor[HTML]{FFFF99}0.9762 \\
M-I-A~\cite{guo2025make} & 1.4098  & 27.7357  & 0.7224  \\
Anymate~\cite{deng2025anymate} & 1.5737  & \cellcolor[HTML]{FAC791}2.6093 & 0.9682  \\
Puppeteer~\cite{song2025puppeteer} & \cellcolor[HTML]{FFFF99}0.5615  & 5.5605 & \cellcolor[HTML]{FAC791}0.9798 \\
Ours & \cellcolor[HTML]{F59194}0.0033 & \cellcolor[HTML]{F59194}1.0453 & \cellcolor[HTML]{F59194}0.9999   \\
\bottomrule
\end{tabular}
}
\vspace{-1.0em}
\label{tab:ours_vs_baselines_unseenobj}
\end{table}

\noindent\textbf{Evaluation Metrics.}
Since our framework is entirely self-supervised and does not rely on annotated skinning data as ground truth, we draw from matrix analysis~\cite{horn1994topics, axler2024linear} and spectral theory~\cite{stoica2005spectral, de2016spectral} to propose three quantitative metrics that measure \emph{structural independence},
\emph{numerical stability}, and \emph{spectral balance} of the skinning weights matrix $W \in \mathbb{R}^{N\times K}$, where $N$ is the number of vertices and $K$ is the number of handles or bones. 
To ensure a fair comparison, we perform $\ell_2$-normalization on each column of \(W\) before evaluation for all methods, resulting in the matrix \(\hat{W}\). 
 
\noindent \textbf{(1) Orthogonality Metric.}
We first measure the pairwise orthogonality among handle or bone influence vectors as:
\begin{equation}
\Omega_{\text{orth}} = 
\frac{1}{K(K-1)} \|\,\hat{W}^\top \hat{W} - I\,\|_2^2,
\end{equation}
where $I$ is the $K\times K$ identity matrix. A smaller $\Omega_{\text{orth}}$ indicates that different handle or bone weights are more decorrelated and capture distinct deformation subspaces. Since each column of $\hat{W}$ is $\ell_2$-normalized, $\Omega_{\text{orth}}$ is bounded within $[0, 1]$, 
and a smaller value indicates that each handle or bone encodes an independent deformation mode.

\noindent \textbf{(2) Log-Condition Number Metric.} 
To evaluate the numerical stability of the skinning basis, we compute the log-scaled condition number~\cite{higham1995condition} of the Gram matrix:
\begin{equation}
\kappa_{\log} = \log_2 \Big( 1 + \frac{\lambda_{\max}(\hat{W}^\top \hat{W})}{\lambda_{\min}(\hat{W}^\top \hat{W})} \Big),
\end{equation}
where $\lambda_{\max}$ and $\lambda_{\min}$ are the largest and smallest eigenvalues of $\hat{W}^\top \hat{W}$. 
$\kappa_{\log}$ lies within $[1, +\infty)$; a smaller value indicates a well-conditioned deformation basis, where small transformations result in stable geometric responses.

\noindent \textbf{(3) Spectrum Entropy Metric.} To assess the balance of spectral energy distribution across skinning weights, we define the spectrum entropy metric~\cite{de2016spectral, stoica2005spectral} as:
\begin{equation}
H_{\text{spec}}  =
- \frac{1}{\log K}\sum_{i=1}^{K}
p_i \log p_i, 
\;
p_i = \frac{\lambda_i}{\sum_j \lambda_j},
\end{equation}
where $\{\lambda_i\}_{i=1}^{K}$ are the eigenvalues of $\hat{W}^\top \hat{W}$. $H_{\text{spec}}$ is bounded within $[0, 1]$, where
higher values indicate more uniform spectral energy and balanced, expressive deformation modes.

\begin{table}[t!]
\centering
\caption{
Quantitative comparison on the ShapeNet dataset~\cite{chang2015shapenet}.
}
\resizebox{0.88\linewidth}{!}{
\tabcolsep 8pt
\footnotesize
\begin{tabular}{lccc}
\toprule
\multicolumn{1}{c}{Method} & \multicolumn{1}{l}{$\Omega_{\text{orth}} \times 10^{-2}$ $\downarrow$ } & \multicolumn{1}{l}{$\kappa_{\log}$ $\downarrow$} & \multicolumn{1}{l}{$H_{\text{spec}}$ $\uparrow$}  \\ 
\midrule
RigNet~\cite{xu2020rignet} & \cellcolor[HTML]{FFFF99}0.5130 & 4.5417  & 0.9648  \\
M-I-A~\cite{guo2025make} & 1.5736 & 27.9573  & 0.7006  \\
Anymate~\cite{deng2025anymate} & 5.3520 & 4.9221   & 0.8858  \\
Puppeteer~\cite{song2025puppeteer} & 1.9528  & \cellcolor[HTML]{FAC791}1.4317  & \cellcolor[HTML]{FFFF99}0.9799 \\
Simplicits~\cite{modi2024simplicits} & \cellcolor[HTML]{FAC791}0.2621 & \cellcolor[HTML]{FFFF99}1.5205 & \cellcolor[HTML]{FAC791}0.9941 \\
Ours & \cellcolor[HTML]{F59194}0.0098 & \cellcolor[HTML]{F59194}1.0460 & \cellcolor[HTML]{F59194} 0.9997  \\
\bottomrule
\end{tabular}
}
\vspace{-1.0em}
\label{tab:ours_vs_baselines_seenobj}
\end{table}

\begin{figure}[!t]
    \centering
    \vspace{0.6em}
    \includegraphics[width=0.98\linewidth]{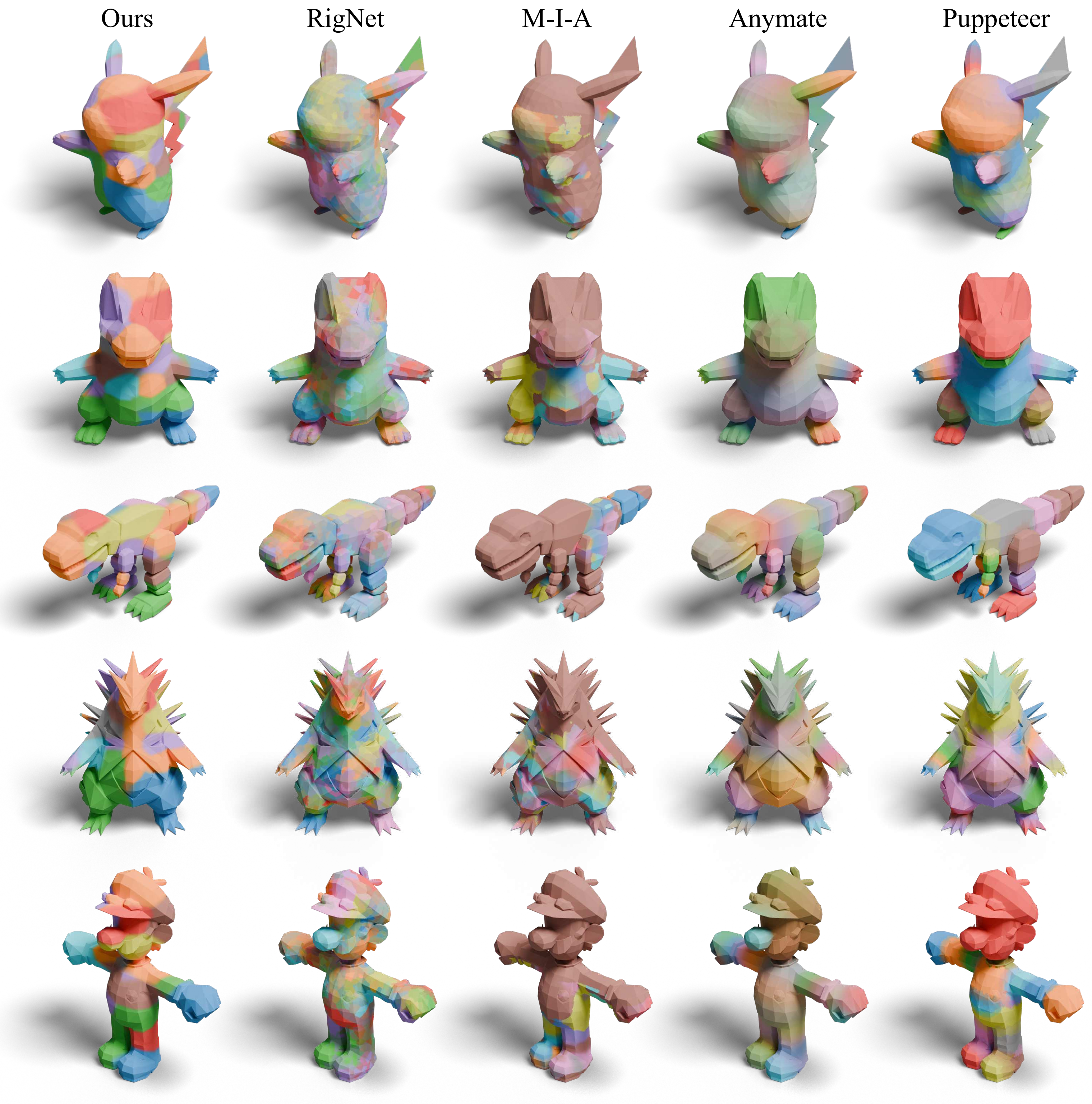}
    \caption{Qualitative comparisons on the RigNet~\cite{xu2020rignet} dataset. We visualize combined skinning fields obtained by blending all skinning weights together, providing an overall view of deformation influences and smoothness compared with baselines~\cite{xu2020rignet, guo2025make, deng2025anymate, song2025puppeteer}.}
    \label{fig:rignet-baselines}
\end{figure}

\begin{figure}[t]
    \centering
    \vspace{0.35em}
    \includegraphics[width=0.98\linewidth]{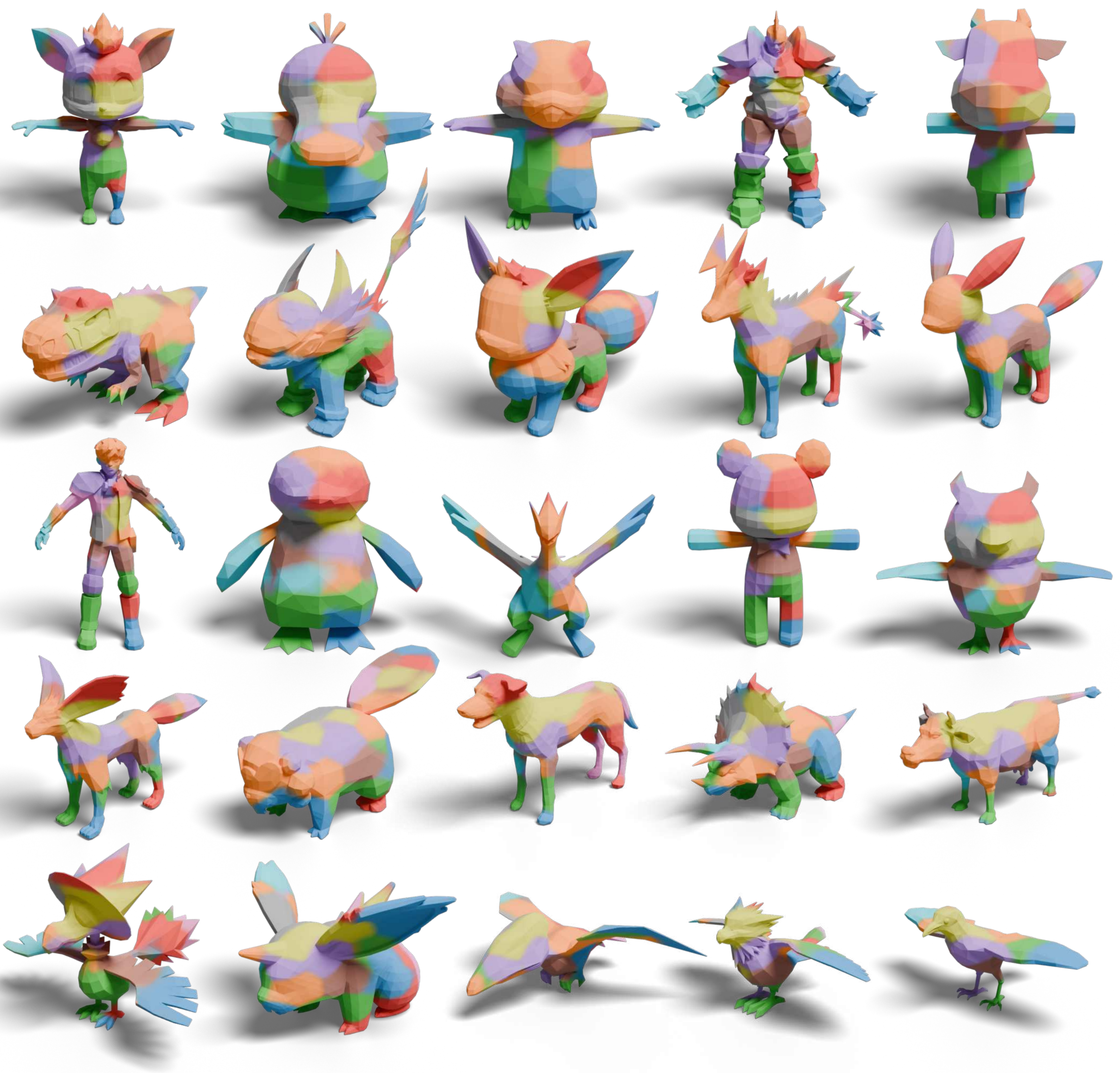}
    \vspace{0.1em}
    \caption{Qualitative results on unseen 3D shapes from the RigNet~\cite{xu2020rignet} test set. We visualize blended skinning fields to demonstrate our method's generalization to novel 3D shapes. \textbf{A single PhysSkin++ model produces all results.}}
    \label{fig:rignet-unseen}
\end{figure}

\begin{figure}[t]
    \centering
    \includegraphics[width=0.98\linewidth]{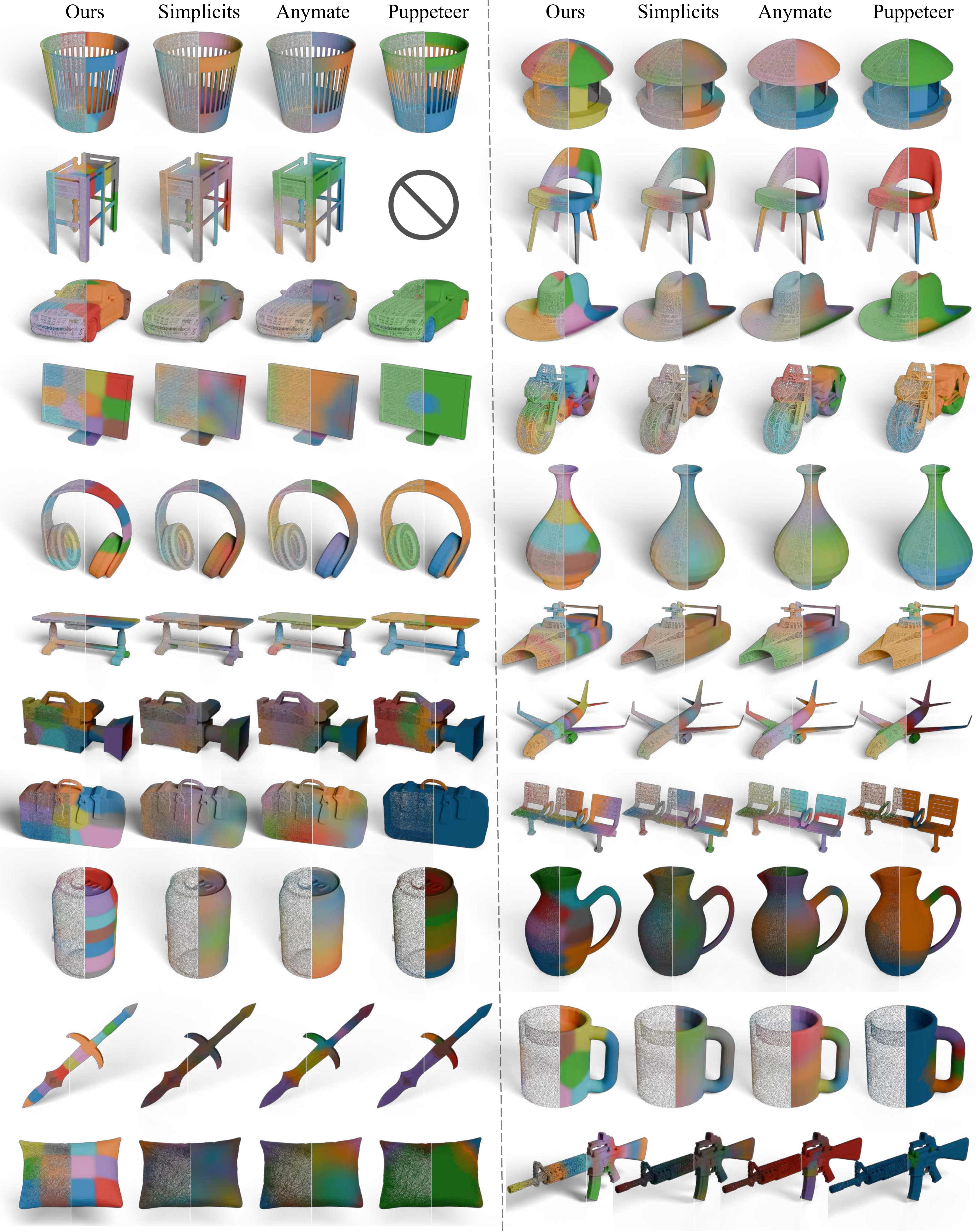}
    \caption{Qualitative comparisons on the ShapeNet~\cite{chang2015shapenet} dataset. We compare with Simplicits~\cite{modi2024simplicits}, Anymate~\cite{deng2025anymate}, and Puppeteer~\cite{song2025puppeteer}. \textbf{Zoom in} for details.}
    \label{fig:ours_vs_baselines_shapenet}
    \vspace{0.3em}
    \includegraphics[width=0.98\linewidth]{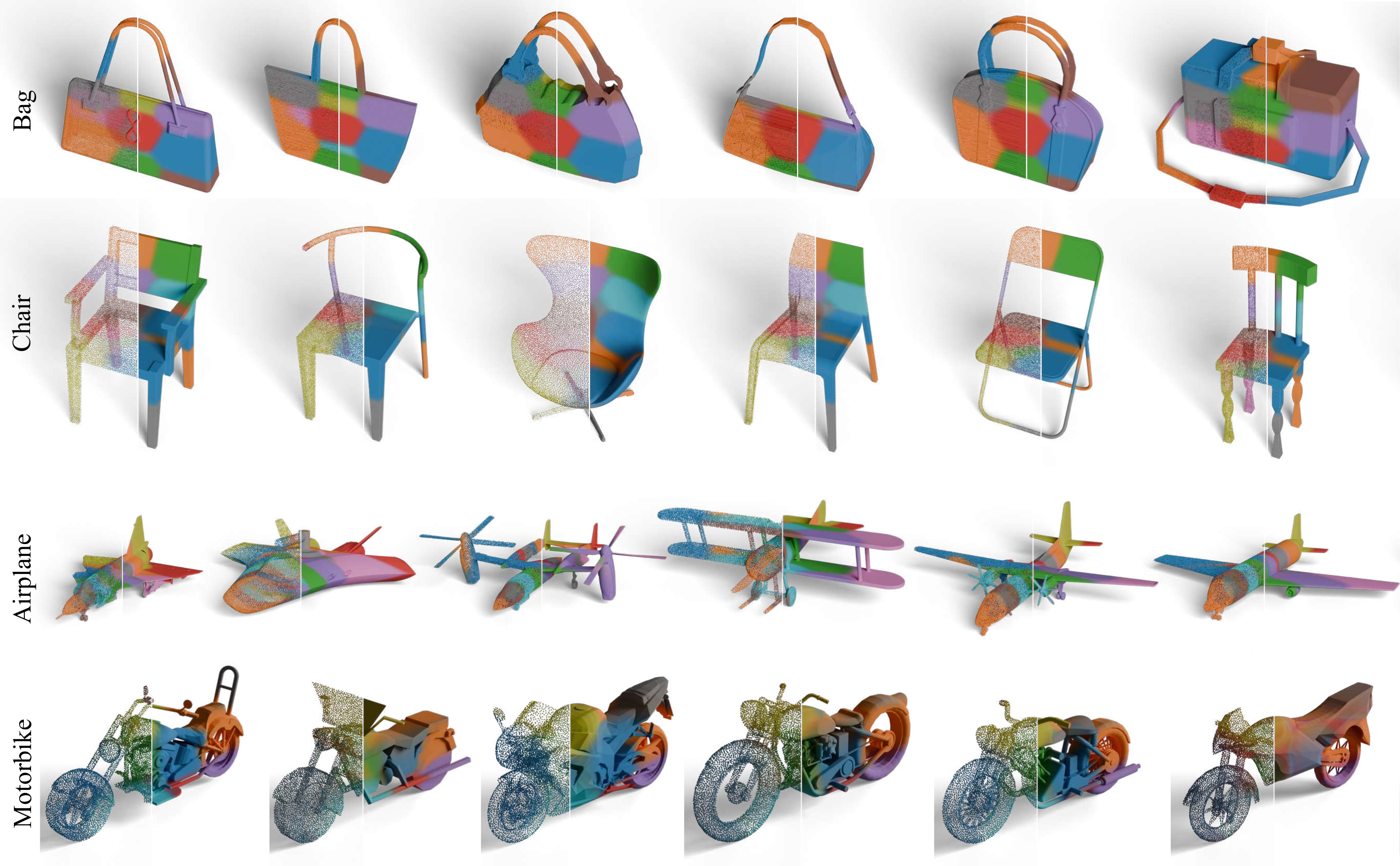}
    \caption{Qualitative skinning results of our method on unseen objects from the ShapeNet~\cite{chang2015shapenet} dataset. \textbf{Zoom in} for more details.}
    \label{fig:ours_unseen}
    \vspace{-0.8em}
\end{figure}

\noindent\textbf{Results.} We first evaluate neural skinning on unseen objects from the test set of the RigNet~\cite{xu2020rignet} dataset to assess generalization. The RigNet dataset provides annotated skinning data; we fine-tune the pretrained feed-forward neural skinning baselines using these annotations, whereas our model is trained only on the 3D meshes. Table~\ref{tab:ours_vs_baselines_unseenobj} reports the quantitative comparison on unseen objects. Our method significantly outperforms the baselines across all three evaluation metrics, demonstrating its generalization ability.
We further evaluate our method on the ShapeNet dataset~\cite{chang2015shapenet}, which provides clearly defined object categories. Our model is trained jointly across all categories, and all baselines are evaluated on the corresponding test sets. Because Simplicits can only be trained on a single object at a time, we train an individual Simplicits model for each object. The quantitative results in Table~\ref{tab:ours_vs_baselines_seenobj} show that our method consistently outperforms all baselines.
In Fig.~\ref{fig:rignet-baselines} and Fig.~\ref{fig:ours_vs_baselines_shapenet}, we compare the qualitative skinning results with the baselines. Our method produces more physically consistent, geometrically orthogonal, and spatially smooth skinning results. Fig.~\ref{fig:rignet-unseen} and Fig.~\ref{fig:ours_unseen} further show that our method generalizes to unseen objects and produces high-quality skinning results. For additional results, please refer to the supplementary material.

\begin{figure}[t]
    \centering
    \includegraphics[width=0.9\linewidth]{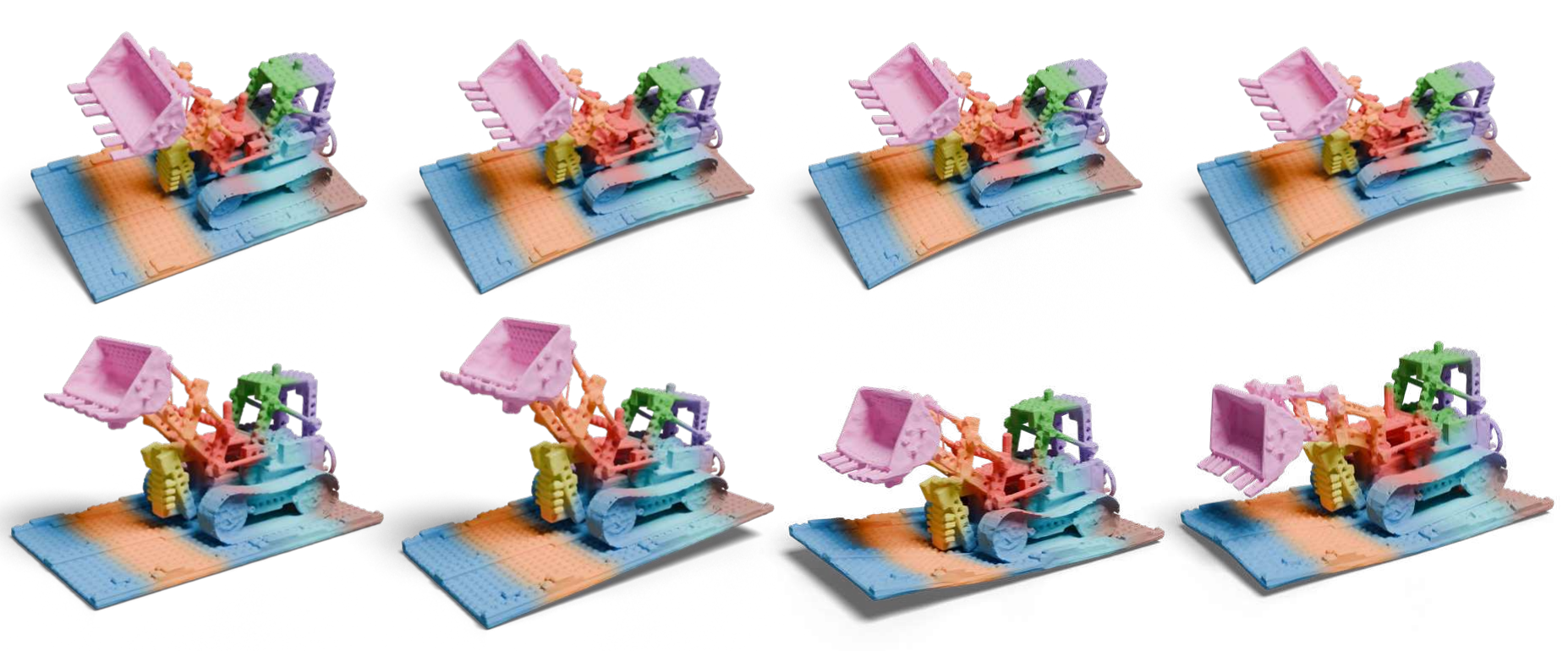}
    \vspace{0.1em}
    \caption{Skinning results for the deformable Lego shape family, where the deformable geometries are from Mani-GS~\cite{gao2024mani}.}
    \label{fig:lego-family}
\end{figure}

\noindent\textbf{Deformable Shape Family.} In Fig.~\ref{fig:lego-family}, our method uses a single model to infer skinning fields for a deformable shape family, while Simplicits~\cite{modi2024simplicits} must be trained separately for each geometry. This comparison demonstrates the efficiency and cross-shape generalization of our framework.

\begin{figure*}[t]
    \centering
    \includegraphics[width=0.95\textwidth]{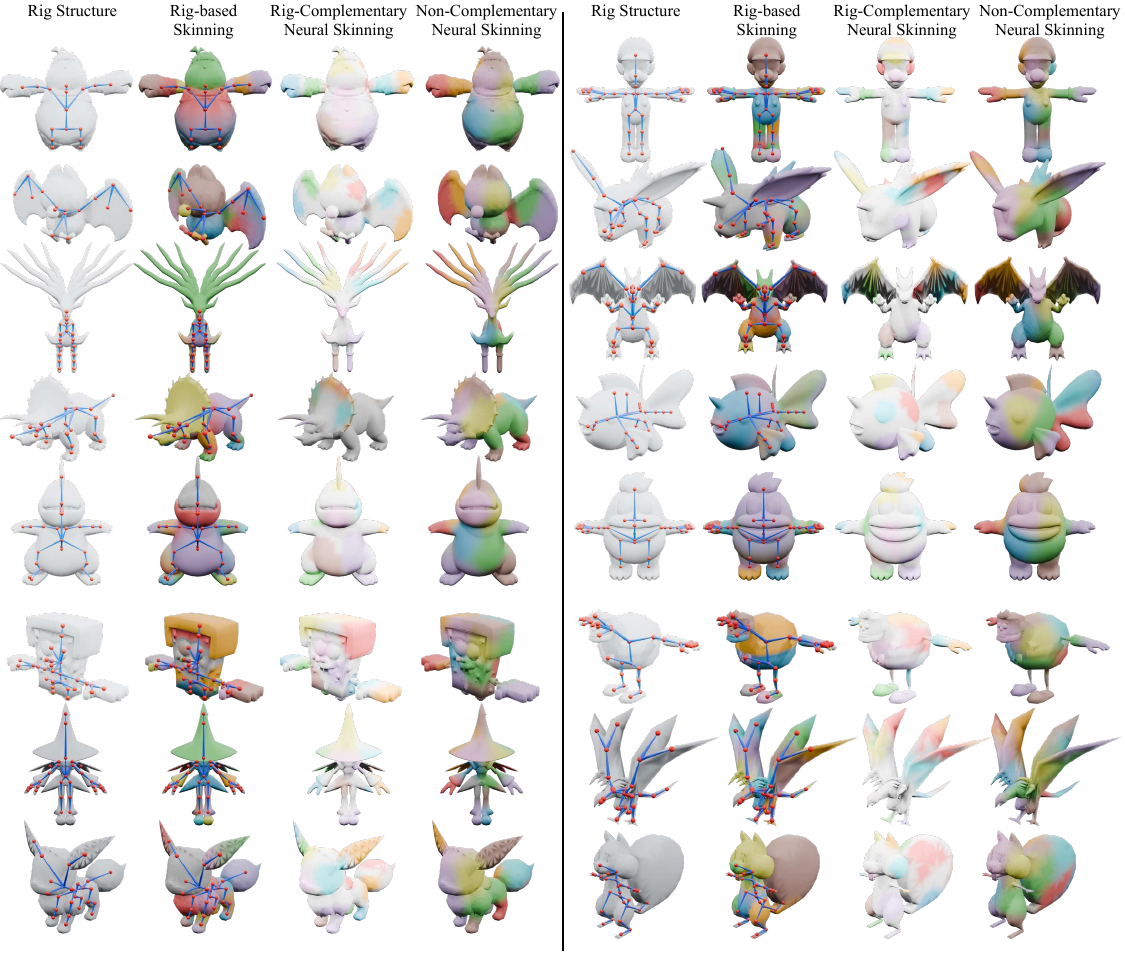}
    \caption{\codex{Qualitative results of self-supervised rig-complementarity learning. For each character, the four columns show the prescribed rig structure, its LBS skinning weights, and neural skinning fields learned with and without the rig-space complementarity term $\mathcal{L}_{\mathrm{comp}}$, respectively. The comparison illustrates how this term changes the spatial distribution of the learned skinning fields across diverse characters and rig structures.}}
    \label{fig:render_comp_skin}
\end{figure*}

\codex{\noindent\textbf{Rig-Complementary Neural Skinning Fields.}
We evaluate the neural skinning fields learned with our self-supervised rig-complementarity objective on characters from the RigNet dataset~\cite{xu2020rignet}, which provides the skeletal rigs and associated LBS skinning weights. Fig.~\ref{fig:render_comp_skin} shows the learned fields across diverse character geometries and rig structures, with spatially varying patterns over the body and appendages. We further examine the resulting complementary elastic responses under interactive rig control and user-authored animation sequences in Sec.~\ref{subsec:animation_experiments}.}

\subsection{Ablation Studies}
\begin{figure}[t]
    \centering
    \includegraphics[width=0.95\linewidth]{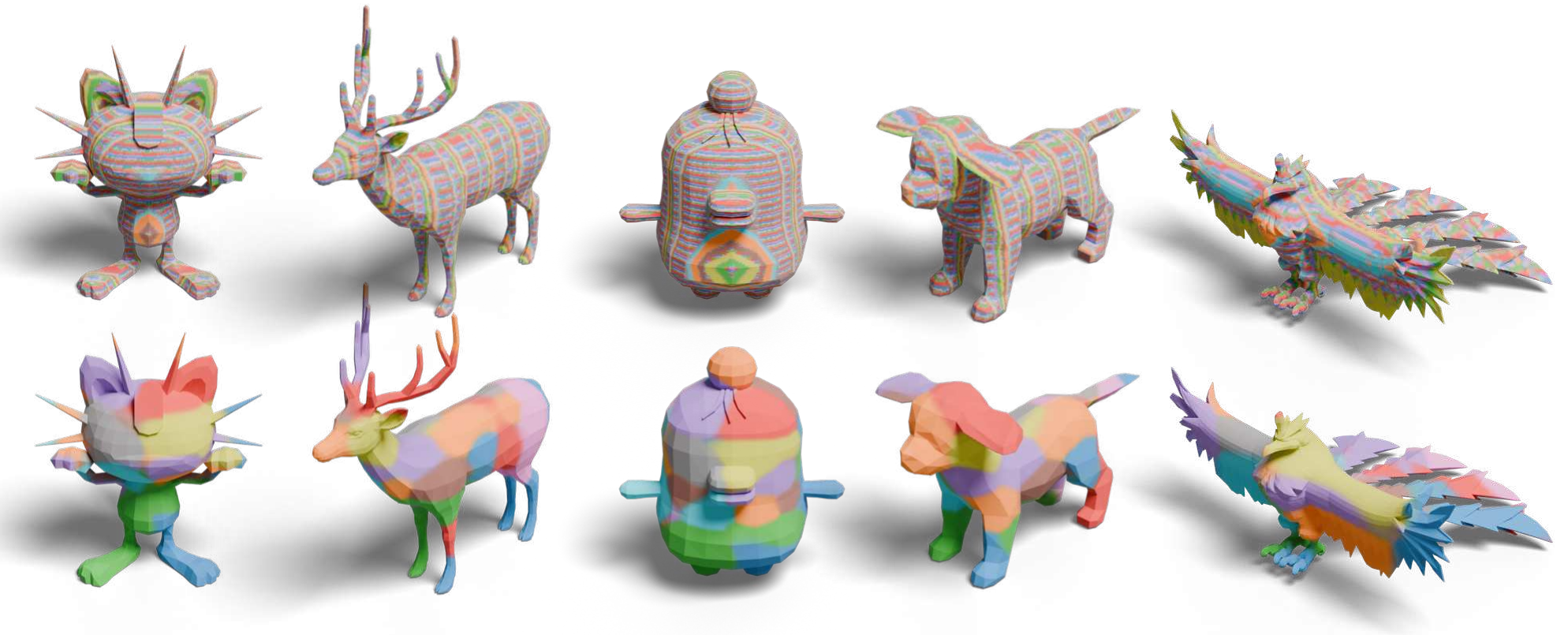}
    \caption{Top: Trained without $\mathcal{L}_{\text{pot}}$. Bottom: Trained with $\mathcal{L}_{\text{pot}}$.}
    \label{fig:ablation_energy}
    \vspace{-0.5em}
\end{figure}

To validate our design choices, we conduct ablation studies covering both the key components of the proposed Physics-Informed Self-Supervised Learning (PISSL) strategy and the rig-space complementarity term in the rig-complementary setting, using the RigNet~\cite{xu2020rignet} dataset. As shown in Table~\ref{tab:ablations}, we remove or modify the PISSL components to assess their individual effects on model performance. Specifically, we evaluate the smoothness loss $\mathcal{L}_{\text{smooth}}$, the orthogonality constraint $\mathcal{L}_{\text{orth}}$, the skinning normalization process, the ONI~\cite{huang2020controllable} layer, and conflict-aware gradient correction using ConFIG~\cite{liu2025config}. The quantitative results demonstrate that these components maintain the numerical stability and orthogonality of the learned neural skinning fields, which are essential for well-conditioned and physically consistent deformation behavior. In Fig.~\ref{fig:ablation_energy}, we visualize the skinning results obtained without the potential energy loss $\mathcal{L}_{\text{pot}}$, where the model fails to produce physically interpretable skinning fields. Fig.~\ref{fig:render_comp_skin} further compares neural skinning fields learned with and without $\mathcal{L}_{\mathrm{comp}}$. Without this term, the visualized fields form broad regions across the head, torso, and limbs. Incorporating $\mathcal{L}_{\mathrm{comp}}$ changes these distributions, with visible differences in the ears, wings, and limb regions across multiple characters. These comparisons illustrate the effect of rig-space complementarity learning on the spatial organization of the skinning fields.

\begin{table}[t]
\centering
\caption{Ablation studies on the RigNet~\cite{xu2020rignet} dataset.}
\label{tab:ablations}
\resizebox{0.95\linewidth}{!}{
\tabcolsep 8pt
\footnotesize
\begin{tabular}{lccc}
\toprule
\multicolumn{1}{c}{Config.} & \multicolumn{1}{l}{$\Omega_{\text{orth}} \times 10^{-2}$ $\downarrow$ } & \multicolumn{1}{l}{$\kappa_{\log}$ $\downarrow$} & \multicolumn{1}{l}{$H_{\text{spec}}$ $\uparrow$}  \\ 
\midrule
$\mathit{w/\!o}$ Skinning Normalization  & 6.5533  & 8.5492  & 0.8113  \\
$\mathit{w/\!o}$ ONI Layer & 0.0081 & 1.0844 & 0.9997     \\
$\mathit{w/\!o}$ ConFIG Optimization & 8.9247 & 11.8595  & 0.7594  \\
$\mathit{w/\!o}$ $\mathcal{L}_{\text{orth}}$ & 100.0  & 29.18  & NaN  \\
$\mathit{w/\!o}$ $\mathcal{L}_{\text{smooth}}$  & 0.0050  & 1.0567  &  0.9998  \\
Full Model & \textbf{0.0033} &  \textbf{1.0453} &  \textbf{0.9999}  \\
\bottomrule
\end{tabular}
}
\vspace{-0.7em}
\end{table}

\subsection{Diverse Real-Time Physics-Based Animation}
\label{subsec:animation_experiments}
\codex{We next examine the broader animation scope enabled by PhysSkin++. Its discretization-agnostic neural skinning fields support efficient reduced dynamics for controllable and collision-aware animation, rig-complementary animation, and direct binding to 3D Gaussian representations.}

\codex{\noindent\textbf{Time Efficiency.} We first evaluate the time efficiency of our skinning-based subspace animation. We compare it with a classical FEM method using implicit time integration and an MPM method using explicit time integration. For these two full-space physical simulators, we use open-source high-performance GPU implementations~\cite{chang2023licrom, zong2023neural}. All methods are evaluated on the same CPU and GPU hardware. As shown in Table~\ref{tab:eval_simtime}, our method demonstrates excellent time efficiency.}

\begin{table}[t]
\centering
\vspace{-0.2em}
\caption{Comparison of per-step simulation cost for physics-based animation with full-space simulators FEM~\cite{chang2023licrom} and MPM~\cite{zong2023neural}.}
\label{tab:eval_simtime}
\resizebox{0.9\linewidth}{!}{
  \setlength{\tabcolsep}{5pt}
  \footnotesize
  \begin{tabular}{lcccc}
    \toprule
    3D Shape &  \shortstack{Vertex\\ Count} & \shortstack{FEM~\cite{chang2023licrom}\\Step Cost (ms)} & \shortstack{MPM~\cite{zong2023neural}\\Step Cost (ms)} & \shortstack{Ours\\Step Cost (ms)} \\
    \midrule
    Airplane & 10K  & 79.83  & 141.83  & \textbf{12.26}   \\
    Bag & 121K & 3012.47  & 233.79   & \textbf{13.39}  \\
    Camera & 80K & 2121.02  & 203.38  & \textbf{12.52}  \\
    Chair & 52K & 1772.70 & 178.46  & \textbf{12.48} \\
    Pillow & 127K & 3170.93  & 251.81  & \textbf{13.74}  \\
    \bottomrule
  \end{tabular}
}
\vspace{-0.2em}
\end{table}

\codex{
\noindent\textbf{Controllable and Collision-Aware Animation.}
Beyond unconstrained deformation, we qualitatively examine how the learned neural skinning subspaces respond to prescribed controls and environmental contact. As shown in Fig.~\ref{fig:controllable_reduced_animation}(a), the Dirichlet boundary conditions enforce zero displacement over the prescribed upper regions, while the unconstrained geometry deforms under gravity. In the interactive sequences of Fig.~\ref{fig:controllable_reduced_animation}(b), the lower region of each object is fixed by Dirichlet boundary conditions, while Gaussian-kernel-weighted dragging provides localized control over the remaining deformable domain. Fig.~\ref{fig:controllable_reduced_animation}(c) further demonstrates contact with planar environments across diverse shapes and boundary orientations, with the contact response modeled by the boundary-penalty energy in Eq.~(\ref{eq:boundary_penalty_energy}).
We further evaluate collision-aware animation involving multiple objects. In the rigid--deformable sequences of Fig.~\ref{fig:multi_object_interaction}(a), a rigid body is released from an elevated position and falls under gravity onto a deformable object. Contact between the two objects is modeled by the point-pair IPC-style barrier in Eq.~(\ref{eq:point_ipc_barrier}), while contact with the supporting floor is handled by the boundary-penalty energy in Eq.~(\ref{eq:boundary_penalty_energy}). In the deformable--deformable sequences of Fig.~\ref{fig:multi_object_interaction}(b), multiple deformable objects are released from elevated positions and fall under gravity. Contacts among the falling objects are resolved using the point-pair IPC-style barrier in Eq.~(\ref{eq:point_ipc_barrier}), while their contact with the supporting floor is modeled by the boundary-penalty energy in Eq.~(\ref{eq:boundary_penalty_energy}). These results demonstrate controllable deformation and multi-object interaction within the PhysSkin++ skinning-based reduced representation.
}

\codex{
\noindent\textbf{Rig-Complementary Animation.}
We evaluate the resulting rig-complementary animation under both interactive control and user-authored animation sequences, as shown in Fig.~\ref{fig:rig_complementary_animation}. For interactive rig control, we implement a Polyscope-based graphical interface~\cite{polyscope} that allows users to manipulate bone poses at runtime. The resulting rig motion defines the primary animation, while PhysSkin++ computes the corresponding complementary elastic response. For user-authored rig animation, we obtain source skeletal animation sequences from Mixamo~\cite{mixamo} and retarget them to the target character rigs using Auto-Rig Pro~\cite{autorigpro}. In both settings, PhysSkin++ augments the primary rig animation with temporally evolving, physics-based complementary deformation.
}

\begin{figure}[!t]
    \centering
    \includegraphics[width=0.95\linewidth]{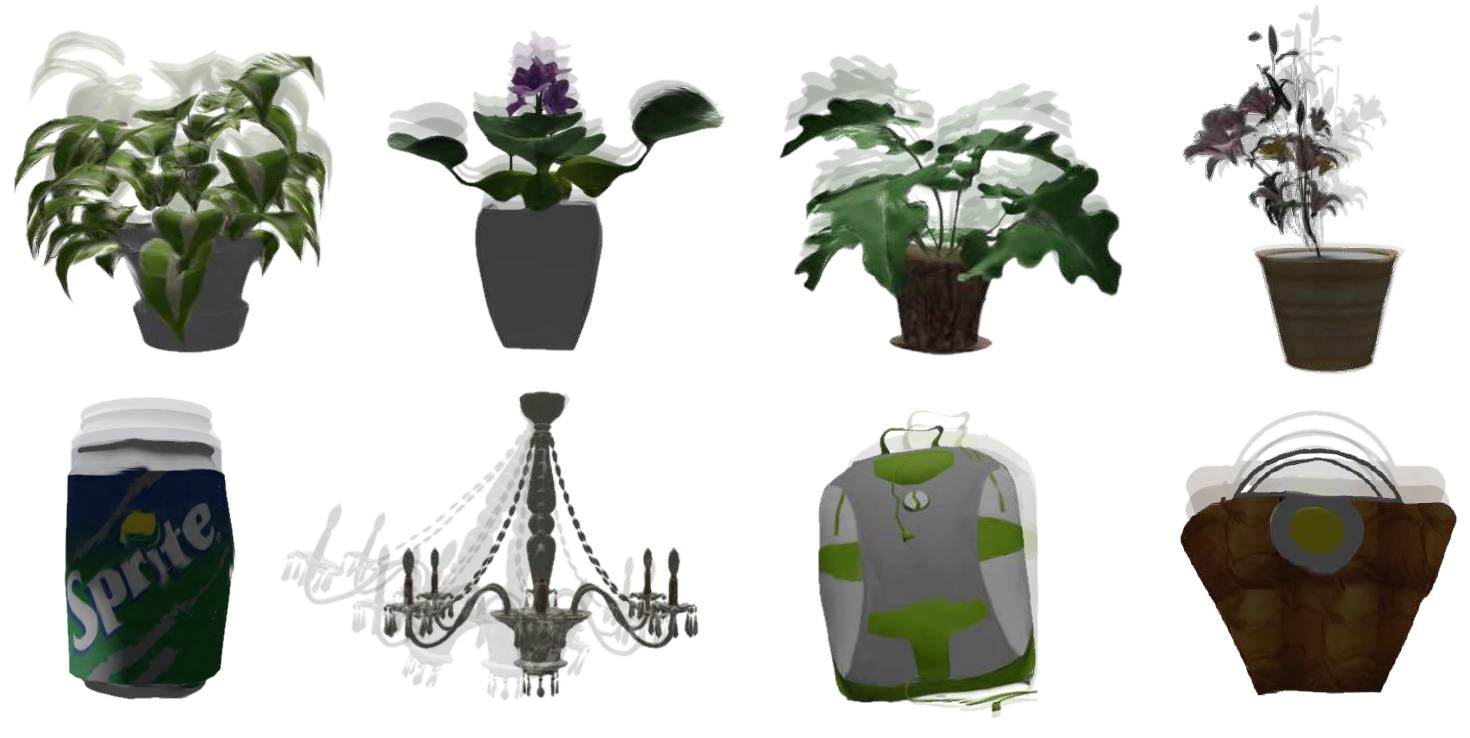}
     \caption{Animation results on 3DGS models~\cite{ma2024shapesplat}.}
    \label{fig:animation-3dgs}
\end{figure}

\codex{\noindent\textbf{Discretization-Agnostic 3D Gaussian Animation.} Owing to the discretization-agnostic nature of our neural skinning formulation, the learned deformation fields can be evaluated directly on 3D Gaussian primitives without introducing an underlying mesh discretization. Following the 3D Gaussian binding formulation in Sec.~\ref{subsec:gaussian_binding}, we evaluate PhysSkin++ on the ShapeSplat dataset~\cite{ma2024shapesplat}, which contains 3D Gaussian reconstructions of objects from ShapeNet~\cite{chang2015shapenet}. As shown in Fig.~\ref{fig:animation-3dgs}, the learned fields are queried at the Gaussian rest centers, and the induced local affine deformation updates both the centers and anisotropic covariances to produce coherent physics-based animation. We encourage readers to visit our project page for additional animation videos.}

\section{Conclusion}
In this paper, we present PhysSkin++, a generalizable physics-informed framework for real-time physics-based animation across diverse 3D shapes and discretizations. A neural skinning fields autoencoder learns continuous skinning fields directly from static geometries to define a skinning-based subspace mapping. Our Physics-Informed Self-Supervised Learning (PISSL) strategy promotes physically plausible deformations through a self-supervised objective and stabilizes joint optimization through on-the-fly skinning-field normalization and conflict-aware gradient correction. At runtime, a reduced dynamics formulation advances the reduced coordinates and supports Dirichlet boundary conditions, interactive control, environmental interaction, and both rigid--deformable and deformable--deformable contact. Beyond purely physics-driven animation, we extend the deformation representation with prescribed rig displacement and the self-supervised objective with a rig-space complementarity term, enabling complementary elastic responses for prescribed rig animations without enforcing the complementarity constraint at runtime. The continuous deformation map also enables discretization-agnostic binding to meshes, point clouds, and 3D Gaussian representations. Experiments demonstrate that PhysSkin++ achieves outstanding performance on generalizable neural skinning while enabling real-time physics-based animation. As a limitation, our current formulation does not incorporate semantic priors for modeling complex geometries. Future work will explore such priors to improve the expressiveness of the learned skinning-based subspace mapping.

\bibliographystyle{paper/ieee_fullname}
\bibliography{paper/egbib,paper/references}

\end{document}